\pdfoutput=1

\documentclass[11pt]{article}

\usepackage[final]{acl}

\usepackage{times}
\usepackage{latexsym}
\usepackage{titletoc}

\usepackage[T1]{fontenc}

\usepackage[utf8]{inputenc}

\usepackage{microtype}

\usepackage{inconsolata}

\usepackage{graphicx}

\usepackage{multirow}
\usepackage{booktabs}
\usepackage{subfigure}
\usepackage{enumitem}
\usepackage{utfsym}
\usepackage{balance}
\usepackage{makecell}
\usepackage{amssymb}
\usepackage{algorithm}
\usepackage{algpseudocode}
\usepackage{diagbox}
\usepackage{colortbl}
\usepackage{xcolor}
\algrenewcommand\algorithmicrequire{\textbf{Input:}}
\algrenewcommand\algorithmicensure{\textbf{Output:}}
\newcommand\blfootnote[1]{%
  \begingroup
  \renewcommand\thefootnote{}\footnote{#1}%
  \addtocounter{footnote}{-1}%
  \endgroup
}

\title{Tailoring Diagnostic Modeling to Individual Learners: Personalized Distractor Generation via MCTS-Guided Reasoning Reconstruction}

\author{
 \textbf{Tao Wu$^1$},
 \textbf{Jingyuan Chen$^{1\dagger}$},
 \textbf{Wang Lin$^1$},
 \textbf{Jian Zhan$^1$},
 \textbf{Mengze Li$^2$},
\\
 \textbf{Fangzhou Jin$^1$},
 \textbf{Min Zhang$^3$},
 \textbf{Kun Kuang$^1$},
 \textbf{Fei Wu$^{1\dagger}$}
\\
 $^1$ Zhejiang University,
 $^2$ Hong Kong University of Science and Technology\\
 $^3$ East China Normal University
\\
\href{mailto:twu22@zju.edu.cn}{\texttt{twu22@zju.edu.cn}},  \ \ \ 
\href{mailto:jingyuanchen@zju.edu.cn}{\texttt{jingyuanchen@zju.edu.cn}}
}

\begin{document}
\maketitle
\blfootnote{$^{\dagger}$ Corresponding authors.}
\begin{abstract}
Distractors—incorrect yet plausible answer choices in multiple-choice questions (MCQs)—are vital in educational assessments, as they help identify student misconceptions by presenting potential reasoning errors.
Current distractor generation methods typically produce shared distractors for all students, ignoring the individual variations in reasoning, which limits their diagnostic effectiveness.
To tackle this challenge, we introduce the task of \textbf{Personalized Distractor Generation}, which tailors distractors to each student's specific cognitive flaws, inferred from their past question-answering (QA) history. 
While promising, this task is particularly demanding due to the limited number of QA records available for each student, which are insufficient for training, as well as the absence of their underlying reasoning process.
To overcome this, we propose a novel, training-free two-stage framework. 
In the first stage, Monte Carlo Tree Search (MCTS) is used to reconstruct the student's reasoning process from past errors, creating a student-specific misconception prototype. 
In the second stage, this prototype guides the simulation of the student's reasoning on new questions, generating personalized distractors that resonate with their individual misconceptions. 
Our experiments, conducted on 1,361 students across 6 subjects, demonstrate that this approach outperforms existing methods in generating plausible, personalized distractors, and also effectively adapts to group-level settings, highlighting its robustness and versatility. 
Project page: \href{https://mccartney01.github.io/pdg}{https://mccartney01.github.io/pdg/}.

\end{abstract}

\section{Introduction}
Multiple-choice questions (MCQs) are widely used in educational assessment, especially in online learning where large-scale evaluation is essential \cite{mcqsurvey1, mcqgen1}. 
A key component of MCQs is the design of \textit{distractors}—incorrect but plausible answer choices intended to uncover student misconceptions by targeting specific errors \cite{dgsurvey1}.

\begin{figure}[t]
\centering
\includegraphics[width=\linewidth]{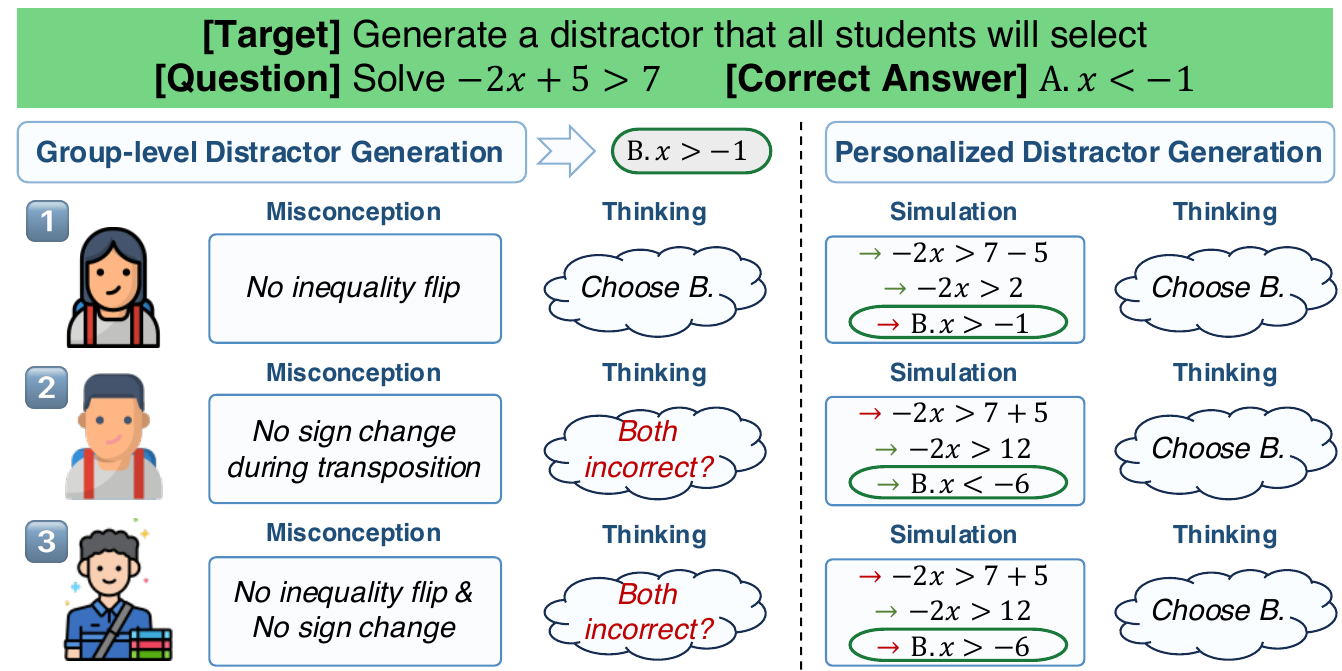} 
\caption{Group-level distractors often fail to reveal diverse student misconceptions, while personalized distractors better align with individual reasoning errors.}
\label{fintro1}
\end{figure}

Recent studies have explored automating distractor generation using large language models (LLMs), typically by learning common error patterns across large student populations \cite{disgem, dgen, dgfirst}.
These methods generate a \textbf{shared} set of distractors per question that reflect broadly observed misconceptions \cite{oanda, divert}. 
However, such \textbf{group-level distractors} often fail to capture the diverse, student-specific nature of reasoning errors \cite{dgsurvey1, person1}. 
When a student's misconception does not align with the predefined distractors, the question loses its diagnostic effectiveness \cite{diageffe}. 
For example, as shown in Figure \ref{fintro1}, a distractor targeting one misconception (\textit{e.g.}, failing to flip the inequality sign) fails for students 2 and 3, whose errors stem from different issues (\textit{e.g.}, incorrect transposition). As a result, they may guess, leaving their misconceptions unexposed and the question diagnostically ineffective.

To address this, we propose a new task, \textbf{Personalized Distractor Generation}, which aims to generate tailored distractors based on each student's specific misconceptions identified from their historical errors\footnote{We provide detailed comparison between personalized and group-level distractor generation in Appendix \ref{acomparison}.}.
Unlike group-level distractors, personalized distractors precisely align with individual misconceptions, enabling more accurate diagnoses across diverse learners, as shown in Figure \ref{fintro1}.

Can existing distractor generation methods tackle the personalized distractor generation task?
Unfortunately, no. 
Existing methods typically rely on large-scale student QA records to learn group-level error patterns, making them ineffective at capturing individual student's unique misconceptions with limited QA records \cite{limiteddata1}. 
Additionally, MCQA records only include selected options without intermediate reasoning steps, making it hard to infer why students chose specific distractors or identify underlying misconceptions \cite{limitdata2}.
A potential solution is to leverage LLMs to reconstruct these missing reasoning processes, given their strong reasoning capabilities \cite{llmreasoning}.
However, this is challenging, as LLMs are typically optimized to produce correct explanations, not to simulate student-like errors \cite{studentsimulation, can}.
Fine-tuning on error-rich data may help, but it risks injecting incorrect knowledge and degrading the model’s overall reasoning performance \cite{sonkar2024llm}.

To tackle these challenges, we propose a training-free two-stage framework for personalized distractor generation.
In the first stage, we analyze each student's past QA records to extract key concepts and reconstruct their reasoning trajectories using a Monte Carlo Tree Search (MCTS) approach.
MCTS breaks down problem-solving into steps, sampling errors at each step to explore potential reasoning paths that may lead to the chosen distractor.
The search is guided by a self-evaluation mechanism that favors realistic student-like reasoning, resulting in the most plausible trajectory.
This trajectory, along with the extracted knowledge concepts, forms a personalized misconception prototype that captures the student's errors.
In the second stage, given a new question, we retrieve relevant misconceptions from the prototype and simulate the student's reasoning process to generate a personalized distractor reflecting the misconceptions.

To support our study, we curate a dataset, \textbf{\texttt{Student\_1361}}, covering 1,361 students across 6 subjects, ranging from elementary to undergraduate levels. 
The dataset provides temporally stable MCQA sequences, making it ideal for personalized distractor generation. 
Experiment results show that our method consistently outperforms strong baselines in generating plausible, student-specific distractors.
Additionally, real-world experiments confirm the diagnostic effectiveness of personalized distractors in revealing individual misconceptions.
Overall, our contributions are as follows:
\begin{itemize}[noitemsep,nolistsep,leftmargin=*]
\item We define the novel task of personalized distractor generation, aimed at producing student-specific distractors based on individual misconceptions to enhance diagnostic effectiveness.

\item We introduce a training-free two-stage framework that uses MCTS to reconstruct student reasoning and generate personalized distractors.

\item Extensive experiments show that our method excels at generating student-specific distractors, with real-world experiments validating the diagnostic effectiveness of personalized distractors.
\end{itemize}

\section{Related Work}
\subsection{Distractor Generation}
Recent work has leveraged large language models (LLMs) to automate plausible distractor generation for MCQs \cite{dgfirst, dgmethod1}, typically by learning common error patterns from large-scale student response data \cite{mcqgen1, disgem}. Early approaches such as option tracing \cite{ot1, ot2} have studied student answer patterns, but they typically formulate the problem as prediction over predefined options or latent knowledge states rather than open-ended distractor generation. More recent works have introduced large language models to move beyond this index-based formulation, enabling open-ended generation of plausible distractors in natural language. For example, D-GEN \cite{dgen} models distractor distributions using MCQ datasets, \citet{oanda} ranks candidates by plausibility, DiVERT \cite{divert} encodes group-level misconceptions, and LOOKALIKE \cite{lookalike} uses direct preference optimization \cite{dpo} to align with typical student errors. However, these methods produce shared, group-level distractors that fail to capture individual reasoning errors, limiting their diagnostic value \cite{dgsurvey1, person1}.
To address this gap, we propose personalized distractor generation, which aims to generate student-specific distractors that reflect each learner’s individual misconceptions and reasoning patterns, thereby enabling more fine-grained and effective diagnostic assessment.

\subsection{Cognitive Diagnosis}
Cognitive diagnosis aims to assess students' mastery of knowledge concepts and identify their misconceptions, providing interpretable insights for personalized learning \cite{cdsurvey, cddong, llmedu}. 
Recent approaches often rely on deep neural networks that encode students' cognitive states as latent parameters \cite{mgncd,hcd}, which, while effective for performance prediction, lack interpretability and make it difficult to pinpoint specific reasoning errors. 
Efforts like student simulation \cite{studentsimulation} or remediation \cite{remediation} attempt explicit modeling of cognitive states in natural language, but require detailed reasoning traces that are rarely available in multiple-choice question settings. 
To address this, we propose an MCTS-based approach that reconstructs reasoning trajectories directly from selected answers, enabling interpretable and student-specific misconception modeling for personalized distractor generation.

\begin{figure*}[t]
\centering
\includegraphics[width=\textwidth]{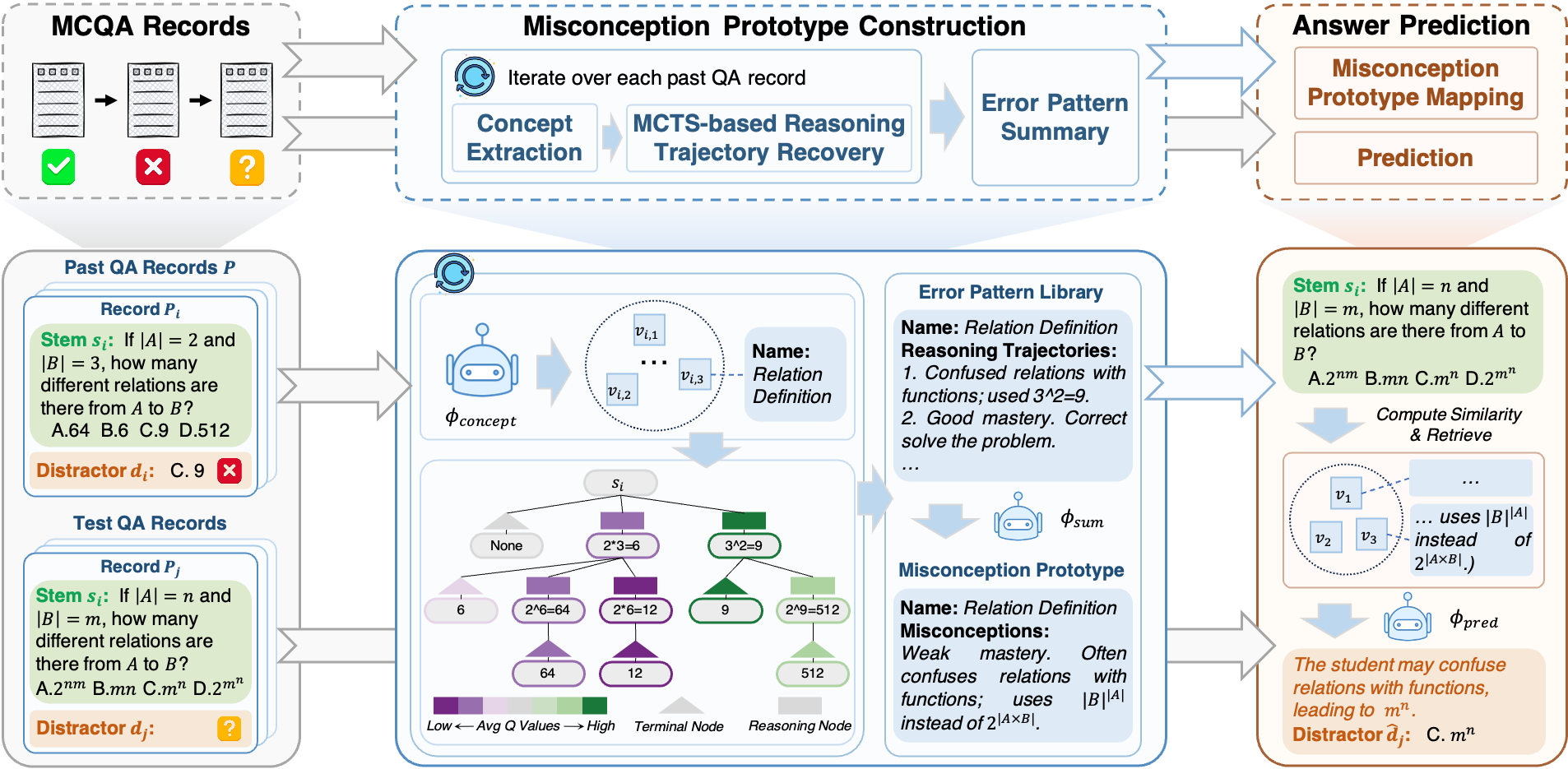} 
\caption{Overview of our two-stage personalized distractor generation framework. In the first stage, given a student's multiple-choice question-answering (MCQA) records, we construct a personalized misconception prototype. This involves: (1) extracting relevant knowledge concepts for each record; (2) applying Monte Carlo Tree Search (MCTS) to reconstruct plausible reasoning trajectories that lead to the student's selected distractors; and (3) summarizing these error trajectories and related concepts into a personalized misconception prototype. In the second stage, given a new question, we retrieve relevant misconceptions from the prototype, feed them into the model to simulate a plausible reasoning, and generate a personalized distractor.}
\label{fmethod}
\end{figure*}

\section{Dataset Curation}
Personalized distractor generation assumes that a student's misconceptions remain stable over short periods, enabling reliable analysis and the generation of individualized distractors. This requires datasets with two essential properties: (1) temporally stable cognitive states for each student and (2) sequential MCQA records.

To meet these requirements, we construct a dataset called \textbf{\texttt{Student\_1361}}, which includes 1,361 students across 6 subjects. These subjects span a wide range of difficulty levels, from elementary-level to undergraduate-level topics, including mathematics, computer science, and more. To ensure diversity, the dataset combines data sourced from publicly available datasets (\textit{e.g.}, Eedi \cite{eedi}) and data collected from an online question-answering platform. These data sources provide a large volume of student answer records and challenging questions, making it ideal for testing and generating personalized distractors.

To ensure quality and consistency, we apply several processing steps. First, we retain only the QA sequences completed within a week for each student to maintain cognitive stability. We also filter out questions unsuitable for distractor generation, such as open-ended or ambiguous question stems (\textit{e.g.}, ``\textit{Which of the following is true?}''). Detailed filtering criteria are provided in Appendix \ref{afilter}.

The final \texttt{Student\_1361} dataset contains a total of around 80K QA records, with each student having an average of 49 historical QA records and 9 test records. The test QA records consist only of questions the student answered incorrectly and are used to evaluate personalized distractor generation methods. The ground-truth distractor is the actual incorrect option selected by the student, serving as a reliable benchmark for this task. Dataset statistics and illustrations are provided in Appendix \ref{adataset}.

\section{Method}
\subsection{Overview}
Personalized distractor generation focuses on creating student-specific incorrect answer options that reflect individual misconceptions, rather than generic distractors shared across large student populations.
To tackle this task, we propose a two-stage framework that operates independently for each student.
In the first stage, given a student's $M$ past QA records $P=\{P_i\}_{1\leq i\leq M}=\{(s_i,a_i,d_i)\}_{1\leq i\leq M}$, where $s_i,a_i,d_i$ denote question stem, correct answer and selected distractor, respectively, we analyze each selected distractor to reconstruct a plausible reasoning process. This helps identify recurring error patterns, which are then summarized into a personalized misconception prototype.
In the second stage, for each new question $s_j$ ($M+1 \leq j \leq M+N$, where $N$ is the number of the student's test QA records), we leverage this prototype to simulate the student’s most likely reasoning process, predict their answer, and generate a tailored distractor that reflects their specific misconceptions.

\subsection{Misconception Prototype Construction}
\label{misprocons}
To generate personalized distractors, it is essential to first understand the student's recurring misconceptions, which are often embedded in erroneous reasoning processes \cite{lau2011guessing}. To achieve this, we construct a personalized misconception prototype for each student. Specifically, for each past QA record, we extract key knowledge concepts and apply Monte Carlo Tree Search (MCTS) \cite{zhang2024rest, zhang2024accessing, tscite} to reconstruct a plausible reasoning trajectory. These trajectories are then summarized into a personalized prototype representing the student's typical misconceptions. This process involves 3 key steps.

\subsubsection{Concept Extraction}
To generate effective personalized distractors, it's crucial to understand where a student's misconception stems from.
In multiple-choice questions, a wrong answer can stem from confusion between multiple concepts.
To support precise and interpretable misconception analysis, we use the model $\phi_{concept}$ to automatically extract relevant concepts for each record $P_i$:
\begin{equation}
    [v_{i,1},v_{i,2},\cdots] = \phi_{concept}(P_i)
\end{equation}
where $[v_{i,1},v_{i,2},\cdots]$ denote the extracted concepts.
This enables the framework to pinpoint errors related to specific concepts, facilitating accurate generation of student-specific distractors.

\subsubsection{MCTS-based Reasoning Trajectory Recovery}
Student misconceptions often manifest in their incorrect reasoning processes, which are crucial to reconstruct in order to identify underlying errors—especially since such reasoning steps are typically absent in MCQA settings.
Current LLMs might struggle to simulate such error-prone reasoning within a single prompt, as they are typically optimized to generate correct answers \cite{studentsimulation}.
To address this, we propose an MCTS-based approach that decomposes the reasoning into multiple steps. 
At each step, the model explores potential errors, and through the search, it identifies the most plausible reasoning trajectory that leads to the student's chosen distractor. 
This trajectory serves as the foundation for analyzing the student's misconceptions.
We then outline the MCTS node definition and its four phases—selection, expansion, simulation, and backpropagation—with pseudocode and a detailed illustration of the whole MCTS process provided in Appendix \ref{amethod}.

\textbf{Node Definition.} 
In our framework, each node in the reasoning tree represents a single reasoning step, with the intermediate result being either correct or incorrect.
However, errors don't always result from explicit mistakes. 
Some students may follow correct reasoning steps but stop prematurely, leading to incorrect conclusions.
For example, as illustrated in Figure~\ref{fmethod}, a student solving a relation-counting problem might correctly compute $2 \times 3 = 6$, but mistakenly stop there, missing the crucial next step of calculating $2^6$.

To account for such behaviors, we define two types of nodes: \textit{reasoning nodes} and \textit{terminal nodes}. 
Each reasoning node represents a step in the reasoning process, followed by a terminal node that marks where the reasoning stops. 
Terminal nodes indicate the end of a reasoning and are not further expanded. 
This design allows variable-length reasoning paths, helping to capture a wide range of student error patterns.

\textbf{Selection.}
The selection phase begins at the root node and recursively selects the best child until reaching one of the conditions: a terminal node, a node that is not fully expanded (defined in the Expansion phase), or the maximum tree depth $D$.

Following the standard MCTS strategy used in board games, each node $n_k$ maintains two statistics: $V$, the number of visits, and $Q$, the accumulated reward from simulations. At each selection, the Upper Confidence Bound for Trees (UCT) formula is applied to choose the best child node:
\begin{equation}
    UCT(n_k)=\frac{n_k.Q}{n_k.V}+c\sqrt{\frac{\log(n.V)}{n_k.V}}
\end{equation}
where $n$ is the parent node of $n_k$ and $c$ is a constant, typically set to $\sqrt{2}$.

\textbf{Expansion.}
The expansion phase extends the selected reasoning node by generating its child nodes. 
Since LLMs can theoretically generate an unlimited number of next steps \cite{chen2024alphamath}, we limit each reasoning node to at most $B$ children ($B \geq 2$), excluding its terminal child, to maintain search efficiency. 
One of these children represents the correct next step, assuming flawless reasoning, while the remaining $B - 1$ are plausible but incorrect alternatives. 
To avoid redundant error steps, all $B - 1$ incorrect children are generated together in a single expansion.
This means all children of a node are generated when it is first expanded. 
Thus, a reasoning node is considered not fully expanded if any of its non-terminal children remains unvisited. 
In such cases, one unvisited child is randomly selected for further simulation.

\textbf{Simulation.}
The simulation phase completes a full reasoning trajectory starting from the selected child node. Given the current partial reasoning path, the model continues to generate subsequent steps until a final answer is reached, simulating how a student might reason from that point. The resulting answer is then passed to the backpropagation phase for trajectory evaluation.

\textbf{Backpropagation.}
The backpropagation phase evaluates and propagates rewards for two types of reasoning trajectories: (1) the full trajectory extended through simulation, and (2) the partial trajectory ending at the selected node without further simulation.
Each trajectory is evaluated based on two criteria. The first is \textit{match score} $r_1$, which indicates whether the simulated final answer matches the selected distractor:
\begin{equation}
    r_1 = 1 \ \ if \ \ d=d_i \ \ else  \ \ 0
\end{equation}
where $d$ is the simulated final answer and $d_i$ is the selected distractor. 
While essential, it is not sufficient on its own, as early in the search, the correct distractor may not yet be reached, yielding zero reward and limited exploration. Additionally, even when the answer matches, the reasoning might still be implausible.

To address this, we introduce the \textit{plausibility score} $r_2\in [0,1]$, which is obtained by prompting the model $\phi_{eval}$ to assess the trajectory's coherence and realism. 
The final reward is a weighted sum: $r=r_1+\alpha r_2$, where $\alpha$ controls the influence of the plausibility score. 
This reward is then propagated backward to update statistics of each node along the trajectory.

After running MCTS for $L$ iterations, we select the most plausible trajectory $t_i$ that leads to the selected distractor.
Combined with the associated knowledge concepts, this trajectory captures both the reasoning path and the underlying misconception, which are stored in the student's error pattern library to form their misconception prototype.

\begin{table*}[ht]
\centering
\resizebox{\linewidth}{!}{
\begin{tabular}{c|ccc|ccc|ccc|ccc|ccc}
\toprule
\multirow{2}[2]{*}{\textbf{Methods}} & \multicolumn{3}{c|}{LLaMA-3-70B} & \multicolumn{3}{c|}{Deepseek-V3} & \multicolumn{3}{c|}{Claude-4-Sonnet} & \multicolumn{3}{c|}{GPT-3.5-turbo} & \multicolumn{3}{c}{GPT-4o} \\
\cmidrule{2-16}
 & Acc & Plaus & Coh & Acc & Plaus & Coh & Acc & Plaus & Coh & Acc & Plaus & Coh & Acc & Plaus & Coh  \\
\midrule\midrule
\rowcolor{gray!20} \multicolumn{16}{c}{\textit{Eedi\_100}} \\
Random & 10.2 & 2.66 & 2.31 & 25.8 & 3.1 & 2.63 & 22.3 & 3.36 & 2.96 & 11.0 & 2.56 & 2.13 & 18.4 & 2.71 & 2.32 \\
Similarity & 12.7 & 2.8 & 2.43 & 24.2 & 3.15 & 2.71 & 22.8 & 3.44 & 3.09 & 12.2 & 2.62 & 2.22 & 22.9 & 2.95 & 2.55 \\
Level & 15.7 & 2.5 & 2.22 & 25.1 & 2.8 & 2.5 & 23.0 & 2.92 & 2.66 & 16.6 & 2.39 & 2.11 & 25.4 & 2.67 & 2.39 \\
Level+Random & 13.7 & 2.43 & 2.17 & 24.8 & 2.75 & 2.44 & 24.0 & 2.88 & 2.63 & 12.9 & 2.16 & 1.9 & 24.7 & 2.65 & 2.34 \\
Level+Similarity & 13.7 & 2.43 & 2.19 & 24.7 & 2.75 & 2.44 & 23.7 & 2.83 & 2.56 & 13.1 & 2.16 & 1.88 & 23.9 & 2.63 & 2.33 \\
Mastery & 14.0 & 2.68 & 2.21 & 25.5 & 3.19 & 2.7 & 23.5 & 3.34 & 2.82 & 15.5 & 2.74 & 2.28 & 25.4 & 3.06 & 2.61 \\
IO & 14.2 & 2.74 & 2.3 & 26.4 & 3.29 & 2.83 & 24.1 & 3.51 & 3.1 & 17.4 & 2.73 & 2.25 & 24.4 & 3.0 & 2.53 \\
CoT & 16.2 & 2.74 & 2.27 & 27.1 & 3.35 & 2.83 & 25.3 & 3.52 & 3.12 & 14.9 & 2.73 & 2.25 & 23.7 & 3.0 & 2.54 \\
\midrule
Ours & \textbf{17.7} & \textbf{2.93} & \textbf{2.45} & \textbf{32.1} & \textbf{3.56} & \textbf{2.98} & \textbf{27.0} & \textbf{3.72} & \textbf{3.31} & \textbf{22.1} & \textbf{3.12} & \textbf{2.53} & \textbf{28.2} & \textbf{3.17} & \textbf{2.64} \\
\midrule\midrule
\rowcolor{gray!20} \multicolumn{16}{c}{\textit{Discrete\_40}} \\
Random & 9.0 & 2.17 & 1.78 & 13.5 & 2.54 & 2.07 & 13.5 & 2.91 & 2.48 & 15.5 & 2.45 & 2.04 & 14.3 & 2.49 & 2.08 \\
Similarity & 11.4 & 2.3 & 1.92 & 16.7 & 2.62 & 2.12 & 15.1 & 2.91 & 2.39 & 15.1 & 2.54 & 2.07 & 12.7 & 2.63 & 2.24 \\
Level & 14.7 & 2.34 & 1.93 & 11.0 & 2.49 & 2.08 & 14.7 & 2.75 & 2.38 & 13.9 & 2.36 & 1.9 & 15.5 & 2.48 & 2.09 \\
Level+Random & 15.1 & 2.35 & 1.96 & 15.5 & 2.58 & 2.18 & 14.7 & 2.67 & 2.25 & 15.5 & 2.36 & 1.98 & 15.1 & 2.51 & 2.16 \\
Level+Similarity & 13.1 & 2.3 & 1.88 & 10.6 & 2.44 & 1.99 & 18.4 & 2.79 & 2.34 & 13.9 & 2.29 & 1.89 & 13.9 & 2.44 & 2.11 \\
Mastery & 17.1 & 2.55 & 2.05 & 22.9 & 2.82 & 2.27 & 26.5 & 2.9 & 2.31 & 18.8 & 2.6 & 2.07 & 17.1 & 2.76 & 2.31 \\
IO & 18.4 & 2.54 & 2.11 & 22.0 & 2.95 & 2.33 & 26.5 & 3.04 & 2.47 & 18.0 & 2.6 & 2.1 & 16.3 & 2.69 & 2.26 \\
CoT & 16.7 & 2.59 & 2.15 & 21.2 & 2.96 & 2.38 & 27.8 & 2.99 & 2.42 & 18.0 & 2.64 & 2.16 & 18.0 & 2.72 & 2.26 \\
\midrule
Ours & \textbf{28.6} & \textbf{2.82} & \textbf{2.33} & \textbf{34.7} & \textbf{3.24} & \textbf{2.69} & \textbf{40.4} & \textbf{3.32} & \textbf{2.76} & \textbf{28.6} & \textbf{2.97} & \textbf{2.38} & \textbf{31.0} & \textbf{3.05} & \textbf{2.52} \\
 \bottomrule
\end{tabular}
}
\vspace{-0.5em}
\caption{Performance comparison between our method and baselines. The best results are \textbf{bolded}.}
\label{tmain}
\vspace{-0.5em}
\end{table*}

\subsubsection{Error Pattern Summary}
By applying MCTS-based reasoning recovery to a student's past QA records, we extract a set of knowledge concepts along with their corresponding erroneous reasoning trajectories. 
For each concept $v_k$, we aggregate its associated trajectories $[t_{k,1},t_{k,2},...]$ and summarize them into a generalized misconception $m_k$ using a summarization model $\phi_{sum}$:
\begin{equation}
    m_k=\phi_{sum}(v_k, [t_{k,1},t_{k,2},...])
\end{equation}
This misconception abstracts the specific context of individual questions, allowing the misconception to be generalized across new problems that involve the same concept. The resulting set of concept–misconception pairs forms a personalized misconception prototype, which guides the generation of targeted distractors for future questions.

\subsection{Answer Prediction}
\label{answerpre}
This stage simulates how the student might incorrectly reason through a new question based on prior misconceptions, generating an erroneous answer that serves as a personalized distractor.
Traditional approaches typically identify similar questions based on question stems and use the errors made on these past questions to predict potential mistakes on new ones.
However, this text-level similarity can be misleading, as questions with similar stems may test entirely different knowledge concepts, resulting in distractors that fail to reflect the student’s true weaknesses.

To overcome this, we shift to a concept-level approach. Instead of relying on question stems, we focus on the student’s misconception prototype, which summarizes reasoning errors based on the knowledge concepts involved. 
Given a new question $s_j$, we first identify its relevant concepts, then retrieve the corresponding misconceptions $[m_{j,1},m_{j,2},...]$ from the prototype. 
These generalized misconceptions are then applied to the new question $s_j$ to simulate a more accurate reasoning:
\begin{equation}
    \hat{d_j} = \phi_{pred}(s_j, a_j, [m_{j,1},m_{j,2},...])
\end{equation}
where $a_j$ denotes the correct answer, $\phi_{pred}$ denotes the model.
The output $\hat{d_j}$ serves as a personalized distractor, reflecting the student's specific reasoning errors. By tailoring distractors to individual misconceptions, this method ensures both their plausibility and diagnostic value.

\begin{table}[t]
\centering
\resizebox{\linewidth}{!}{
\begin{tabular}{c|cccccccc}
\toprule
Parameter & \makecell[c]{$M$\\(avg)} & \makecell[c]{$N$\\(avg)} & $D$ & $B$ & $c$ & \makecell[c]{$L$\\(Eedi\_100)} & \makecell[c]{$L$\\(Discrete\_40)} & $\alpha$ \\ \midrule
Value & 49 & 9 & 5 & 3 & $\sqrt{2}$ & 10 & 20 & 0.2 \\
 \bottomrule
\end{tabular}
}
\caption{Key parameter settings of our method.}
\label{tparameter}
\end{table}

\section{Experiments}
\subsection{Experiment Setup}
\subsubsection{Datasets}
Given the substantial computational and financial costs for running extensive experiments on the \texttt{Student\_1361} dataset—with around 100 different experimental settings, as discussed in Section \ref{textmainexp} and \ref{sindepth}—we conduct our analytical experiments on two subsets: \texttt{Eedi\_100}, which contains 100 students focusing on elementary math, and \texttt{Discrete\_40}, which covers 40 undergraduate students working on discrete mathematics. These subsets allow for comprehensive testing while ensuring feasibility. For the main comparison, we extend our analysis to \texttt{Student\_1361}, where we compare our method with the second-best baseline, ensuring that our results are both robust and representative as discussed in Section \ref{sindepth}.

\begin{table*}[ht]
\centering
\resizebox{\linewidth}{!}{
\begin{tabular}{c|ccc|ccc|ccc|ccc|ccc}
\toprule
\multirow{2}[2]{*}{\textbf{Methods}} & \multicolumn{3}{c|}{LLaMA-3-70B} & \multicolumn{3}{c|}{Deepseek-V3} & \multicolumn{3}{c|}{Claude-4-Sonnet} & \multicolumn{3}{c|}{GPT-3.5-turbo} & \multicolumn{3}{c}{GPT-4o} \\
\cmidrule{2-16}
 & Acc & Plaus & Coh & Acc & Plaus & Coh & Acc & Plaus & Coh & Acc & Plaus & Coh & Acc & Plaus & Coh  \\
\midrule\midrule
\rowcolor{gray!20} \multicolumn{16}{c}{\textit{Eedi\_100}} \\
Full Method & 17.7 & 2.93 & 2.45 & 32.1 & 3.56 & 2.98 & 27.0 & 3.72 & 3.31 & 22.1 & 3.12 & 2.53 & 28.2 & 3.17 & 2.64 \\
w/o $\phi_{concept}$ & 14.3 & 2.86 & 2.34 & 30.9 & 3.53 & 2.96 & 22.9 & 3.58 & 3.06 & 14.4 & 2.89 & 2.36 & 22.8 & 3.04 & 2.51 \\
w/o Terminal Node & 17.4 & 2.9 & 2.44 & 28.1 & 3.51 & 2.89 & 26.1 & 3.68 & 3.22 & 17.4 & 3.06 & 2.45 & 25.8 & 3.1 & 2.57 \\
w/o $\phi_{eval}$ & 16.6 & 2.89 & 2.4 & 27.9 & 3.49 & 2.86 & 26.3 & 3.67 & 3.27 & 16.8 & 3.02 & 2.49 & 25.7 & 3.11 & 2.59 \\
w/o $\phi_{sum}$ & 17.3 & 2.91 & 2.39 & 28.9 & 3.49 & 2.87 & 26.5 & 3.66 & 3.17 & 20.6 & 3.1 & 2.48 & 27.1 & 3.14 & 2.6 \\
\midrule\midrule
\rowcolor{gray!20} \multicolumn{16}{c}{\textit{Discrete\_40}} \\
Full Method & 28.6 & 2.82 & 2.33 & 34.7 & 3.24 & 2.69 & 40.4 & 3.32 & 2.76 & 28.6 & 2.97 & 2.38 & 31.0 & 3.05 & 2.52 \\
w/o $\phi_{concept}$ & 19.6 & 2.69 & 2.23 & 22.4 & 3.06 & 2.43 & 26.5 & 3.15 & 2.51 & 19.6 & 2.79 & 2.23 & 20.4 & 2.96 & 2.35 \\
w/o Terminal Node & 25.7 & 2.71 & 2.31 & 30.2 & 3.14 & 2.54 & 38.8 & 3.18 & 2.6 & 26.5 & 2.81 & 2.35 & 28.2 & 2.97 & 2.47 \\
w/o $\phi_{eval}$ & 22.4 & 2.71 & 2.23 & 31.0 & 3.16 & 2.56 & 31.8 & 3.09 & 2.58 & 24.5 & 2.88 & 2.28 & 24.1 & 2.9 & 2.38 \\
w/o $\phi_{sum}$ & 26.1 & 2.72 & 2.3 & 30.2 & 3.18 & 2.6 & 36.7 & 3.25 & 2.64 & 27.8 & 2.92 & 2.38 & 29.8 & 3.01 & 2.49 \\
 \bottomrule
\end{tabular}
}
\caption{Ablation results on \texttt{Eedi\_100} and \texttt{Discrete\_40} dataset across five LLMs.}
\label{tabl}
\end{table*}

\subsubsection{Parameter Settings}
We list key parameter settings of our framework in Table \ref{tparameter}. 
Here, $M$ and $N$ denote the average number of past and test QA records per student, respectively. 
The number of MCTS iterations $L$ is set based on subject complexity, with reasoning-intensive datasets using larger $L$ values. 
We provide a detailed analysis of this choice in Section~\ref{sindepth}, and report ablation studies on other parameters in Appendix \ref{aexp}.

To ensure model-agnostic evaluation, we use the same underlying LLM for all components of our framework (\textit{e.g.}, $\phi_{concept}$, $\phi_{sum}$, etc.). We experiment with five representative models covering different architectures and scales: two open-source models (\textit{LLaMA-3-70B} \cite{llama} and \textit{DeepSeek-V3} \cite{liu2024deepseek}) and three close-source ones (\textit{Claude-4-Sonnet} \cite{claude}, \textit{GPT-3.5-turbo} \cite{gpt35}, and \textit{GPT-4o} \cite{achiam2023gpt}).

\subsubsection{Baselines}
We compare our method with six heuristic baselines that do not model student reasoning trajectories and two reasoning-aware baselines that attempt to reconstruct student thinking. 
Heuristic baselines include:
1) \textit{Random}: randomly selects a past QA record as reference;
2) \textit{Similarity}: retrieves the most similar record based on stem embeddings;
3) \textit{Level} \cite{level}: estimates the student’s cognitive level from past QA accuracy;
4) \textit{Level+Random}: combines Level-based estimation with random reference selection;
5) \textit{Level+Similarity}: combines Level-based estimation with similarity-based retrieval;
6) \textit{Mastery} \cite{studentsimulation}: models the student’s conceptual mastery but ignores misconceptions.
Reasoning-aware baselines include:
7) \textit{IO} \cite{tot}: directly simulates possible student error trajectories;
8) \textit{CoT} \cite{cot}: uses a chain-of-thought prompting strategy to emulate reasoning.

\subsubsection{Metrics}
We evaluate performance using three metrics. The first is Accuracy (\textit{Acc}), which measures whether the generated distractor matches the student’s selected one. \footnote{We use \texttt{math-verify} \cite{math_verify} to compare generated distractors with the student selected ones.}
The other two are LLM-based metrics that assess the quality of the generated reasoning trajectories: Plausibility (\textit{Plaus}), which reflects how realistic the reasoning is as a student response, and Coherence (\textit{Coh}), which measures its logical consistency and fluency.
Both Plaus and Coh are scored on a 1–5 scale (higher is better), with the advanced o3-mini model serving as the evaluator.

Additional experimental settings and results are provided in Appendix \ref{aexp}, including further evidence that our method can also generate high-quality group-level distractors.

\subsection{Performance Comparison}
\label{textmainexp}
As shown in Table \ref{tmain}, our method consistently outperforms all baselines across both datasets and model backbones, achieving significant improvements in all three metrics. These results highlight the effectiveness of our approach in generating plausible personalized distractors that align with realistic student misconceptions.
Additionally, more powerful LLMs, such as Claude-4-Sonnet, lead to better performance, indicating that stronger models provide richer search spaces and more coherent reasoning candidates during the MCTS process.

Beyond performance improvements, the results also reveal two innovative findings:

1) \textbf{Explicit reasoning trajectory recovery is essential.}
Baselines that simulate reasoning trajectories (\textit{e.g.}, IO, CoT) outperform heuristic methods in most cases, emphasizing the importance of modeling how students arrive at incorrect answers. This is especially critical in MCQ settings, where the selected choice itself offers limited diagnostic value.

2) \textbf{MCTS excels in reasoning-intensive scenarios.}
The advantage of our method is especially prominent on \texttt{Discrete\_40}, where questions require more complex reasoning. While single-pass methods struggle to recover realistic trajectories, MCTS’s iterative exploration enables more accurate reconstruction of plausible misconceptions.

\begin{table}[t]
\centering
\resizebox{\linewidth}{!}{
\begin{tabular}{c|ccc|ccc|ccc}
\toprule
\multirow{2}[2]{*}{\textbf{Subsets}} & \multicolumn{3}{c|}{Eedi\_700} & \multicolumn{3}{c|}{Discrete\_106} & \multicolumn{3}{c}{Python\_192} \\
\cmidrule{2-10}
 & Acc & Plaus & Coh & Acc & Plaus & Coh & Acc & Plaus & Coh \\
 \midrule
 CoT & 15.6 & 3.16 & 2.61 & 20.4 & 2.6 & 2.18 & 17.5 & 2.45 & 2.29 \\
 Ours & 21.0 & 3.47 & 2.9 & 26.5 & 3.23 & 2.64 & 20.3 & 2.69 & 2.39 \\
 \midrule\midrule
 \multirow{2}[2]{*}{\textbf{Subsets}} & \multicolumn{3}{c|}{Java\_200} & \multicolumn{3}{c|}{CP\_39} & \multicolumn{3}{c}{OS\_124} \\
\cmidrule{2-10}
 & Acc & Plaus & Coh & Acc & Plaus & Coh & Acc & Plaus & Coh \\
 \midrule
 CoT & 14.9 & 2.72 & 2.18 & 3.6 & 2.64 & 2.52 & 11.0 & 2.64 & 2.06 \\
 Ours & 17.5 & 2.77 & 2.21 & 6.7 & 2.78 & 2.67 & 12.4 & 2.68 & 2.16 \\
 \bottomrule
\end{tabular}
}
\caption{Performance comparison on \texttt{Student\_1361}.}
\label{twhole}
\end{table}

\subsection{In-Depth Analysis}
\label{sindepth}
We further validate four vital issues as follows.

\subsubsection{Extended Evaluation Across \texttt{Student\_1361}}
We further evaluate our approach on the entire \texttt{Student\_1361} dataset, comparing it with the second-best baseline, CoT. For this comparison, we use the GPT-3.5-Turbo model, with $L=10$ for the \texttt{Eedi\_700} subset and $L=20$ for all other subsets. As shown in Table \ref{twhole}, our method outperforms CoT across all subjects, demonstrating the robustness and generalizability of our approach.

\subsubsection{Effectiveness of Each Component}
We perform ablation studies to evaluate the contribution of each core component in our two-stage framework. 
1) Replacing the concept extractor $\phi_{concept}$ with simple stem-level similarity causes a substantial drop in performance. This is likely due to the limitation of surface-level retrieval in math problems, where similar wording often masks different underlying concepts.
This confirms \textit{the necessity of concept-level reasoning for accurate misconception analysis.}
2) Removing terminal nodes disables the modeling of truncated reasoning paths—common in real student errors where reasoning halts prematurely. The drop in performance demonstrates their value in capturing diverse student reasoning errors.
This validates \textit{the importance of explicitly modeling early termination.}
3) Eliminating the plausibility signal $\phi_{eval}$ from the reward function leaves only sparse answer-matching supervision, especially ineffective during early search stages.
This underscores \textit{the role of plausibility in shaping effective search guidance.}
4) Feeding raw reasoning trajectories into the prediction model leads to degraded performance, likely due to information overload and redundancy from key misconceptions.
This shows \textit{the benefit of distilling error patterns via trajectory summarization.}

\begin{figure}[t]
\centering
\includegraphics[width=\linewidth]{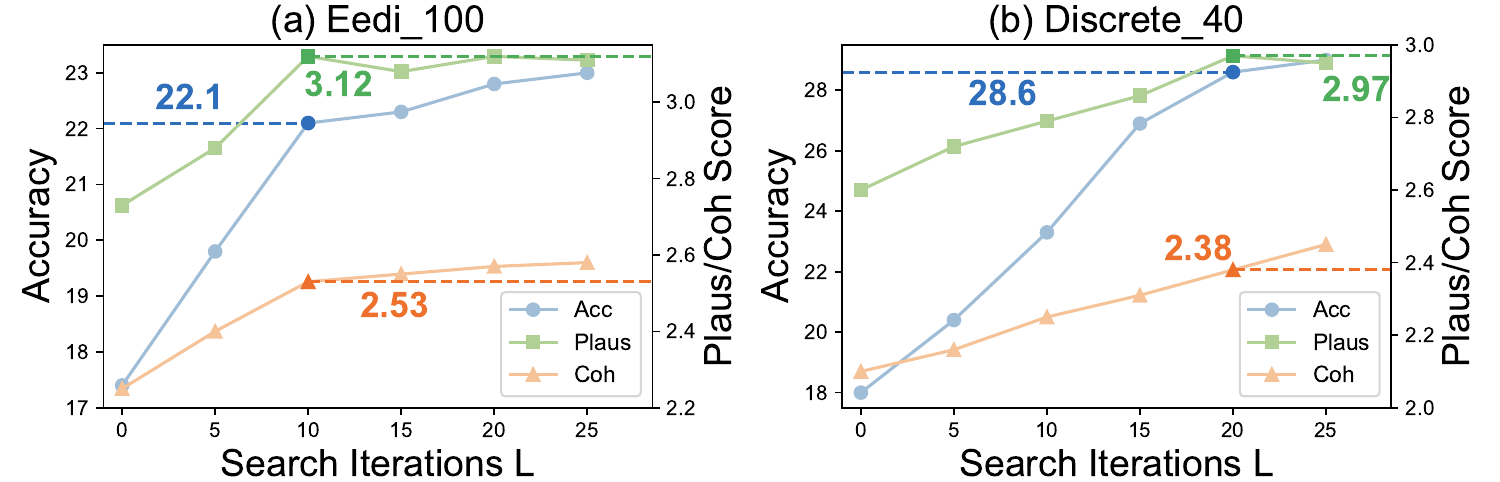} 
\caption{Results of different MCTS search iterations.}
\label{fsearch}
\end{figure}

\subsubsection{Impact of MCTS Search Iterations}
We examine the impact of varying MCTS search iterations $L$ using GPT-3.5-turbo, as shown in Figure~\ref{fsearch}.
On the \texttt{Eedi\_100} dataset, all evaluation metrics improve up to $L{=}10$, after which performance plateaus or slightly declines, indicating that shallow searches are sufficient for simple problems and deeper exploration may introduce noise.
In contrast, on the more challenging \texttt{Discrete\_40} dataset, performance continues to improve steadily up to $L{=}20$, demonstrating the need for deeper exploration in reasoning-intensive tasks.
These trends support our default choice of $L$ values for different datasets.

\begin{table*}[t]
\centering
\resizebox{0.72\linewidth}{!}{
\begin{tabular}{c|cc|cc|cc|cc|cc}
\toprule
\multirow{2}[2]{*}{Methods} & \multicolumn{2}{c|}{LLaMA-3-70B} & \multicolumn{2}{c|}{Deepseek-V3} & \multicolumn{2}{c|}{Claude-4-Sonnet} & \multicolumn{2}{c|}{GPT-3.5-turbo} & \multicolumn{2}{c}{GPT-4o} \\
\cmidrule{2-11}
 & Plaus & Coh & Plaus & Coh & Plaus & Coh & Plaus & Coh & Plaus & Coh\\
 \midrule\midrule
 CoT & 2.54 & 2.13 & 2.89 & 2.27 & 2.85 & 2.39 & 2.64 & 2.12 & 2.72 & 2.25 \\
 Ours & 2.77 & 2.32 & 3.16 & 2.59 & 3.25 & 2.64 & 2.81 & 2.23 & 2.88 & 2.43 \\
 
 \bottomrule
\end{tabular}
}
\vspace{-0.5em}
\caption{Human evaluation of the generated distractors on \texttt{Discrete\_40}.}
\label{thuman}
\end{table*}

\subsubsection{Human Evaluation}
We further conduct a human evaluation study on \texttt{Discrete\_40} to assess the plausibility of the distractors generated by our method. 
Considering the large number of experimental settings and the cost of manual evaluation, we compare our method with the strongest competing baseline, CoT, which is the second-best-performing method according to the LLM-based evaluation results in Table \ref{tmain}. 
This choice enables a focused comparison against a strong reference model.

We recruit 20 undergraduate students with solid backgrounds in Discrete Mathematics as evaluators. 
Following the same evaluation criteria as the LLM-based setting, they rate each simulated solution in terms of \textit{Plaus} and \textit{Coh}. 
Each distractor is independently evaluated by two raters, and the final score is computed as the average of their ratings.

As shown in Table \ref{thuman}, our method consistently receives higher human evaluation scores than CoT across all model settings. This alignment between human judgments and LLM-based evaluation provides evidence that the distractors generated by our method are more plausible and coherent.

\subsection{User Study and Case Analysis}
To empirically evaluate the effectiveness of personalized distractors, we conduct a user study with 30 students sampled from \texttt{Discrete\_40}.
These students are required to answer new questions selected from a held-out dataset, CEval Discrete Math \cite{ceval}.
As shown in Table \ref{tuser}, the students are divided into two groups with comparable error rates on \texttt{Discrete\_40}: 
a control group, which receives original CEval questions with default distractors, 
and a treatment group, which receives modified distractors, including one personalized distractor (generated specifically for each student by our method using Claude-4-Sonnet) and two frequent distractors (selected from the most common predicted distractors across all students).

To better assess whether the distractors truly reflect students' misconceptions, we add an additional choice to each question: ``\textit{None of the above}''. Selecting this option means none of the distractors reflect the student's actual misconceptions, indicating insufficient diagnostic effectiveness.

\begin{table}[t]
\centering
\resizebox{\linewidth}{!}{
\begin{tabular}{c|cccc}
\toprule
Group & \makecell[c]{Error Rate\\(Discrete\_40)} & \makecell[c]{Error Rate\\(CEval)} & Abstain Rate & PD Rate \\
\midrule\midrule
Control & 0.22 & 0.12 & 0.14 & - \\
Treatment & 0.23 & 0.21 & 0.06 & 0.4\\
 \bottomrule
\end{tabular}
}
\caption{Results of the user study, comparing original CEval distractors for the control group with personalized distractors for the treatment group.}
\label{tuser}
\vspace{-0.5em}
\end{table}

\begin{figure}[t]
\centering
\includegraphics[width=\linewidth]{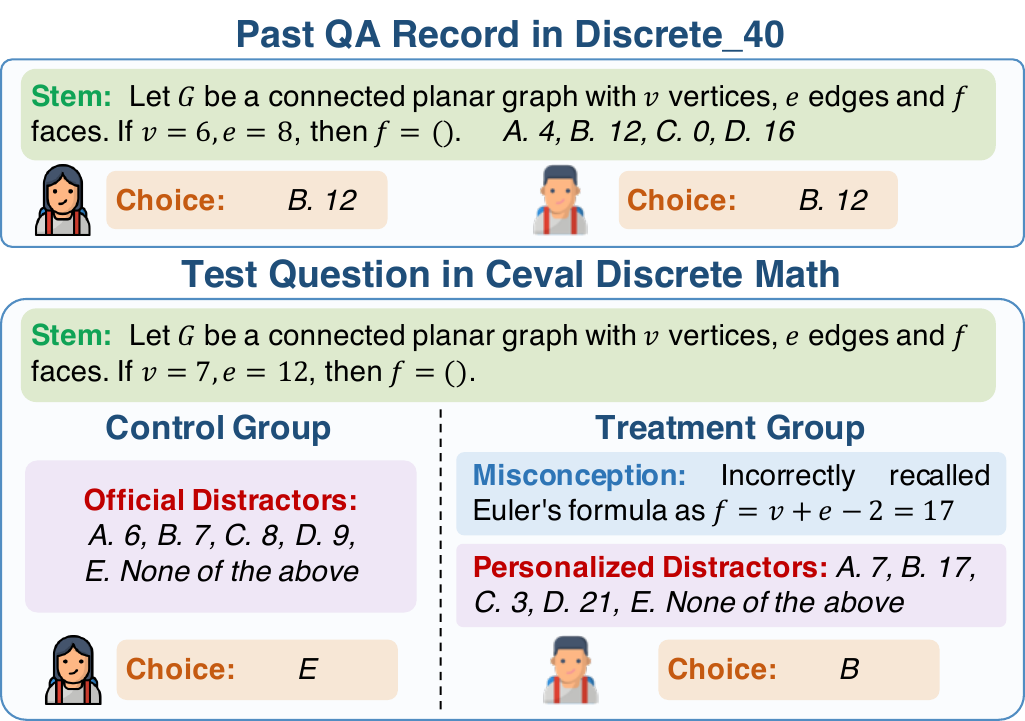} 
\caption{The control-group student chooses ``\textit{None of the above}'', because none of the default group-level distractors reflects their actual misconception. In contrast, the treatment-group student selects the personalized distractor that accurately reflects the misconception.}
\label{fcase}
\vspace{-0.5em}
\end{figure}

We evaluate the performance using three metrics:
1) \textit{Error Rate}: the proportion of responses selecting a distractor (excluding the correct answer and ``\textit{None of the above}'');
2) \textit{Abstain Rate}: the proportion of responses selecting ``\textit{None of the above}'';
3) \textit{PD Rate}: among all incorrect responses, the proportion that selects the personalized distractor.

Results in Table \ref{tuser} show that the treatment group exhibits a higher error rate, lower abstain rate, and substantial PD rate, suggesting that personalized distractors effectively align with students’ misconceptions and improve diagnostic effectiveness.

Figure~\ref{fcase} presents a representative case for this effect. Two students had previously misunderstood Euler's formula and selected the same incorrect answer. When presented with a similar question, the control group student selects ``\textit{None of the above}'', as none of the default distractors matches the misconception. In contrast, the treatment group student chooses the personalized distractor, which accurately reflects his prior misunderstanding, demonstrating the advantage of personalization in targeting specific cognitive errors.

\section{Conclusion}
We propose the novel task of personalized distractor generation to precisely diagnose student misconceptions. 
To address sparse QA data and missing reasoning traces, we introduce a training-free MCTS-based framework that reconstructs student reasoning and generates tailored distractors. 
Experiments show that our method produces more accurate distractors, and the user study demonstrates the diagnostic value of personalized distractors.

\clearpage

\section*{Limitations}
In this section, we discuss the limitations of our work as follow:
\begin{itemize}[noitemsep,nolistsep,leftmargin=*]
    \item Our framework relies on past QA records to support reliable personalization, and performance may be less strongly individualized for students with extremely sparse data. However, the system does not break down in such cases; it naturally falls back to concept-level or group-level distractors and improves as more records accumulate.
    \item The reasoning trajectories reconstructed by LLM-guided MCTS might not exactly replicate students’ true cognitive processes; rather, they represent plausible approximations inferred from their past responses. Although such approximations may occasionally diverge from the student’s actual reasoning at a fine-grained level, our experiments show that the generated distractors remain pedagogically meaningful and effective for diagnosing misconceptions. Future work may further improve faithfulness by incorporating richer behavioral signals or confidence-aware filtering when constructing misconception prototypes.
    \item Although our framework involves multiple model components (\textit{e.g.}, different $\phi$-modules) and may appear structurally complex, it remains computationally and financially feasible in practice. Many modules operate in lightweight or shared-inference settings, several stages can be parallelized, and the system scales smoothly under realistic workloads. We provide detailed analyses in Appendix \ref{appcost} and \ref{apperror}.
\end{itemize}

\section*{Ethics Statement}
\texttt{Student\_1361} uses only pre-existing or anonymized data sources. The \texttt{Eedi\_700} subset is sourced from the publicly available Eedi dataset and contains no personally identifiable information. The other 5 subsets are constructed from data on an online platform. To ensure privacy, we remove all sensitive information and retain only anonymized student IDs. Additionally, we confirm that all user-generated data is strictly authorized under the platform's \textit{terms of service} during user registration process. 

The user study was reviewed and approved by an independent institutional ethics committee prior to data collection. All participants were adults and provided informed consent before taking part in the study, and participation was voluntary with the option to withdraw at any time. No personally identifiable information was collected, and all data were anonymized and used solely for research purposes.

\section*{Acknowledgements}
This research was supported by grants from the “Pioneer” and ”Leading Goose” R\&D Program of Zhejiang under Grant No. 2025C02022 and the National Natural Science Foundation of China (No.62307032).

\bibliography{main}

\appendix
In this appendix, we present the following content:

\startcontents
\printcontents{}{1}{}

\section{Detailed Comparison Between Group-level and Personalized Distractor Generation}
\label{acomparison}
We summarize the main difference between group-level and personalized distractor generation as follows:

\begin{figure*}[t]
\centering
\includegraphics[width=\linewidth]{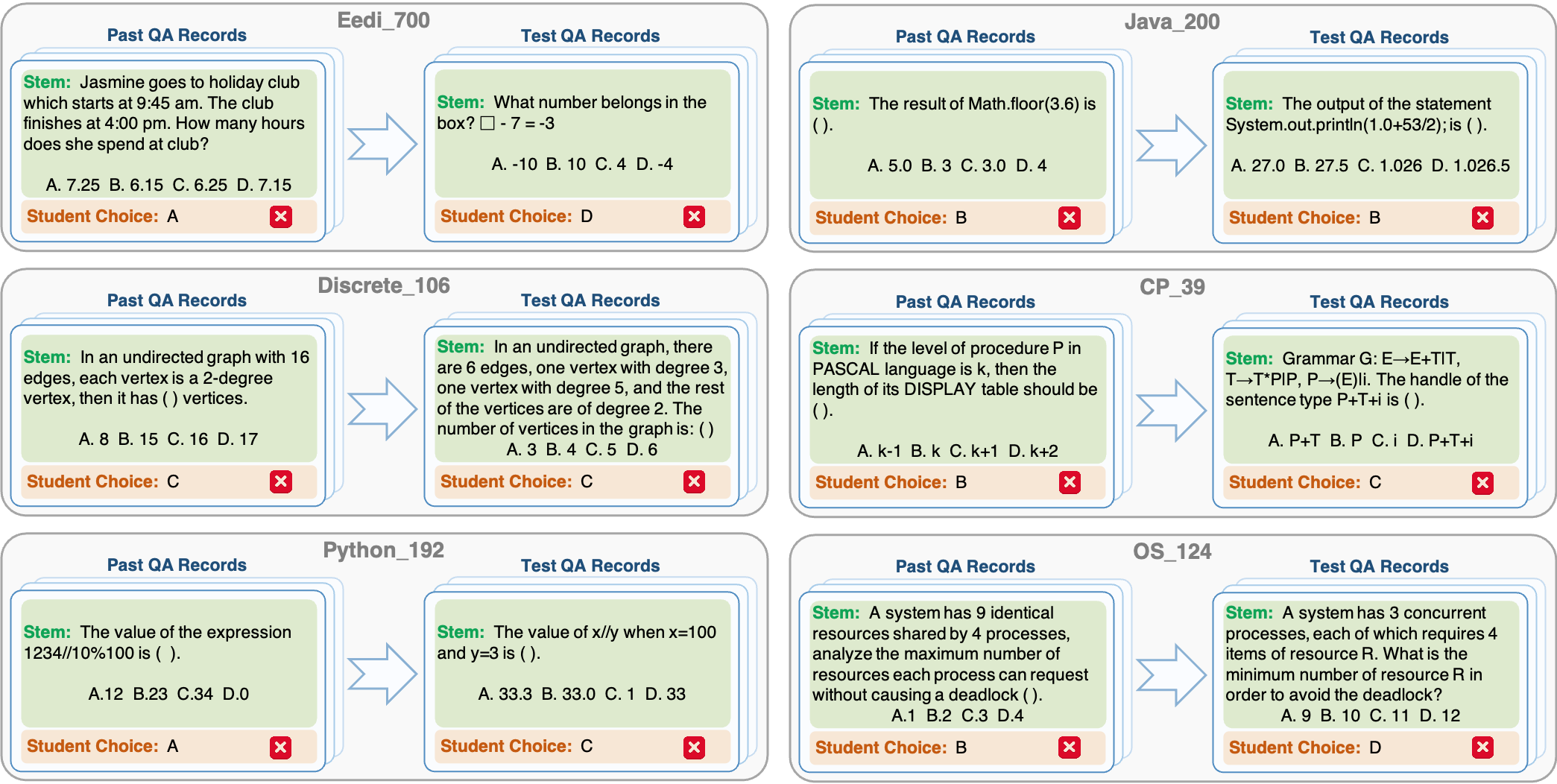} 
\caption{Sample illustrations of \texttt{Student\_1361}.}
\label{fadataset}
\end{figure*}

\begin{table*}[t]
\centering
\resizebox{\linewidth}{!}{
\begin{tabular}{c|cccccc}
\toprule
Subset & Subject & Source & Difficulty & Student Number & $M$ (avg) & $N$ (avg) \\
\midrule\midrule
Eedi\_700 & Math & Eedi \cite{eedi} & Elementary & 700 & 63 & 10 \\
Discrete\_106 & Discrete Math & Online Platform & Undergraduate & 106 & 37 & 10 \\
Python\_192 & Python Programming & Online Platform & Undergraduate & 192 & 47 & 10 \\
Java\_200 & Java Programming & Online Platform &  Undergraduate & 200 & 32 & 10\\
CP\_39 & Compiler Principles & Online Platform & Undergraduate & 39 & 17 & 5 \\
OS\_124 & Operating System & Online Platform & Undergraduate & 124 & 24 & 5 \\
 \bottomrule
\end{tabular}
}
\caption{Statistics of \texttt{Student\_1361}.}
\label{tstatistics}
\end{table*}

\begin{itemize}
    \item \textbf{Objective and application scenario.} Group-level distractor generation aims to produce generic distractors applicable to a wide range of students. It is typically used in standardized testing or large-scale problem database construction. In contrast, personalized distractor generation focuses on generating student-specific incorrect options that reflect each learner’s unique misconceptions. This makes it especially suitable for online learning \cite{chen2024wisdombot} and adaptive assessment \cite{styleadapt} settings.
    \item \textbf{Data requirements and input format.}
Group-level distractor generation typically requires datasets with only the question stem and a fixed set of distractors. Some methods also leverage statistical signals from answer distributions, such as the proportion of students selecting each option \cite{oanda}.
Personalized distractor generation, on the other hand, requires a richer input: a sequence of past QA records for each student, including both the question stem and the specific distractor selected. This enables fine-grained modeling of individual cognitive patterns.

\item \textbf{Generation strategy and cognitive modeling.}
Group-level distractor generation often relies on extracting frequent error patterns from a large population, but it lacks the ability to capture individual differences in reasoning.
Personalized distractor generation emphasizes reconstructing how each student reasons incorrectly, often by modeling their error trajectories and constructing a personalized misconception prototype to guide generation.

\item \textbf{Evaluation methodology and objectivity.}
In group-level distractor generation, datasets usually provide several pre-defined distractors per question, and evaluation is performed by measuring how many of the generated distractors match the ground truth \cite{dgfirst}. However, due to the diversity of student error types, fixed ground-truth distractors cannot cover all plausible mistakes. This makes the evaluation less objective. Some recent work proposes LLM-based evaluation system \cite{qgsms}, but it incurs high cost.
In contrast, personalized distractor generation has a clear and objective evaluation criterion: if the generated distractor matches the student’s actual selected incorrect answer, it is considered correct. This offers a more direct and interpretable evaluation of distractor quality.

\end{itemize}

\section{Dataset Details}
\label{adataset}
\subsection{Dataset Statistics and Sample Illustrations}
Our \texttt{Student\_1361} dataset covers a wide range of subjects and cognitive levels, spanning from elementary mathematics to multiple undergraduate domains, including discrete mathematics, Python programming, Java programming, compiler principles, and operating systems. 
In total, it contains 1,361 students, with each subset providing substantial QA records (on average 50 past records and 10 test records per student). 
This broad subject coverage, variation in difficulty, and large-scale student participation make the dataset highly diverse and representative, providing a rich foundation for studying personalized misconceptions and distractor generation. 
Detailed dataset statistics are reported in Table \ref{tstatistics}, and Figure \ref{fadataset} provides sample examples from each subset.

\subsection{Invalid Data Criteria}
\label{afilter}
To ensure the suitability of each QA record for the distractor generation task, we apply strict filtering rules to exclude questions from the original data source that cannot support reasoning-based analysis or distractor construction. Specifically, a QA record is retained only if it meets the following conditions:

\begin{figure}[t]
\centering
\includegraphics[width=0.7\linewidth]{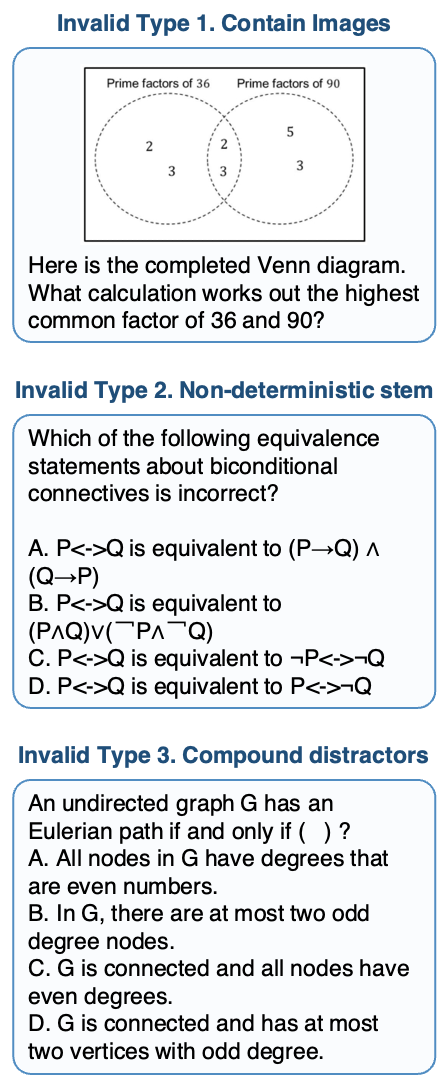} 
\caption{Illustrations of invalid data.}
\label{finvalid}
\end{figure}

\begin{enumerate}
    \item The question stem must be purely textual, without any diagrams or images, ensuring that the problem can be solved based solely on the text. We will further explore extending the framework to multimodal learning scenarios \cite{semanticalignment, autogeo} in future work.

\item The question must be deterministic, meaning the correct answer should be clearly inferable from the stem alone, without requiring the answer choices for disambiguation.

\item All answer choices must be numerical values, formulas, mathematical expressions or code. This restriction ensures objective and accurate evaluation, as existing tools like math-verify perform reliably only on purely mathematical content, while there is no robust method for comparing distractors that contain both natural language and math expressions.
\end{enumerate}

We provide several invalid QA record samples in Figure~\ref{finvalid}.

\section{Method Details}
\label{amethod}
\subsection{Pseudocode of the MCTS algorithm}
we provide pseudocode for the overall MCTS process (Algorithm \ref{algmain}), selection (Algorithm \ref{algselect}), expansion (Algorithm \ref{algexpand}), simulation (Algorithm \ref{algsim}), backpropagation (Algorithm \ref{algback}), function of judging whether a node is fully expanded (Algorithm \ref{algfully}), and function of collecting all the terminal node that match the student's actual chosen distractor after the MCTSeach (Algorithm \ref{algmatch}).

\begin{algorithm}[!t]
    \caption{MCTSearch for reasoning trajectory recovery}
    \label{algmain}
    \begin{algorithmic}[1]
        \Require Question stem $s$, student chosen distractor $d$, max search iteration $L$, max tree depth $D$, max children number $B$, model for generating correct next step $\phi_{correct}$, model for generating several incorrect next steps $\phi_{incorrect}$, model for simulation $\phi_{simulation}$, model for producing plausibility reward $\phi_{eval}$, plausibility reward weight $\alpha$, UCT constant $c$
        \Ensure A plausible reasoning trajectory $t$ that leads to $d$
        
        \State $n_0 \gets create\_node(depth=0,t=None,$
        \Statex        $r=None,is\_terminal=False,Q=0,V=0,child=None,parent=None)$
        \State $l \gets 1$
        \While{$l \leq L$}
            \State $n_l \gets Selection(n_0,D,c)$
            \State $n_l \gets Expansion(n_l,D,B,\phi_{correct},$
            \Statex \hspace{\algorithmicindent} $\phi_{incorrect})$
            \State $t_l,d_l \gets Simulation(n_l,\phi_{simulation})$
            \State $Backpropagation(n_l,\phi_{eval},t_l,d_l,d,\alpha)$
        \EndWhile
        \State Initial empty list of terminal nodes that match the student chosen distractor $MNs = []$
        \State $MNs \gets search\_match\_nodes(n_0,d,$
        \Statex $MNs)$
        \If{$MNs$ has a node}
            \State $n^\ast \gets \max(\frac{\hat{n}.Q}{\hat{n}.V})$ for $\hat{n}$ in $MNs$
        \Else
            \State $n^\ast \gets n_0$
            \While{$n^\ast$ has a child}
                \State $n^\ast \gets \max(\frac{\hat{n}.Q}{\hat{n}.V})$ for $\hat{n}$  in $n^\ast.child$
            \EndWhile
        \EndIf
        \State \Return $t \gets n^\ast.t$
    \end{algorithmic}
\end{algorithm}

\begin{algorithm}[!t]
    \caption{Selection}
    \label{algselect}
    \begin{algorithmic}[1]
        \Require Root node $n_0$, max tree depth $D$, UCT constant $c$
        \Ensure Selected not-fully-expanded node $n_l$
        
        \State $n_l \gets n_0$
        \While{$is\_fully\_expanded(n_l)$ and $n_l.depth < D$}
            \State $n_l \gets \max(\frac{\hat{n}.Q}{\hat{n}.V}+c\sqrt{\frac{\log (n_l.V)}{\hat{n}.V}})$ for $\hat{n}$ in 
            \Statex \hspace{\algorithmicindent} $n_l.child$
        \EndWhile
        \State \Return $n_l$
    \end{algorithmic}
\end{algorithm}

\begin{algorithm}[!t]
    \caption{Expansion}
    \label{algexpand}
    \begin{algorithmic}[1]
        \Require Selected node $n_l$, max tree depth $D$, max children number $B$, model for generating correct next step $\phi_{correct}$, model for generating several incorrect next steps $\phi_{incorrect}$, model for producing plausibility reward $\phi_{eval}$
        \Ensure Expanded node $n_l$
        
        \If{$n_l.depth \ge D$}
            \State \Return $n_l$
        \EndIf

        \If{number of $n_l.child = 0$}
            \State terminal node $\hat{n} \gets create\_node($            
            \Statex \hspace{\algorithmicindent} $depth=n_l.depth+1,t=n_l.t,$
            \Statex \hspace{\algorithmicindent} $r=n_l.r,is\_terminal=True,$
            \Statex \hspace{\algorithmicindent} $Q=0,V=0,child=None,$
            \Statex \hspace{\algorithmicindent} $parent=n_l)$
            \State $n_l.child.append(\hat{n})$
            \State $Backpropagation(n_l.child[0],\phi_{eval},n_l.t,$
            \Statex \hspace{\algorithmicindent}$n_l.r,d,\alpha$)
            
            \State $t_{correct}, r_{correct} \sim \phi_{correct}(n_l)$
            \State correct node $\hat{n} \gets create\_node(depth=$
            \Statex \hspace{\algorithmicindent} $n_l.depth+1,t=[n_l.t,t_{correct}],r=$
            \Statex \hspace{\algorithmicindent} $r_{correct},is\_terminal=False,Q=$
            \Statex \hspace{\algorithmicindent} $0,V=0,child=None,parent=n_l)$
            \State $n_l.child.append(\hat{n})$
            
            \State $t_{incorrect}^1, r_{incorrect}^1,\dots,t_{incorrect}^{B-1},$
            \Statex \hspace{\algorithmicindent} $r_{incorrect}^{B-1} \sim \phi_{incorrect}(n_l)$
            \For{each $k \in [1,2,\dots,B-1]$}
                \State incorrect node $\hat{n} \gets create\_node($
                \Statex \hspace{\algorithmicindent} \hspace{\algorithmicindent} $depth=n_l.depth+1,$
                \Statex \hspace{\algorithmicindent} \hspace{\algorithmicindent}$t=[n_l.t,t_{incorrect}^{k}], r=r_{incorrect}^{k},$
                \Statex \hspace{\algorithmicindent} \hspace{\algorithmicindent} $is\_terminal=False,Q=0,$
                \Statex \hspace{\algorithmicindent} \hspace{\algorithmicindent} $V=0,child=None,$
                \Statex \hspace{\algorithmicindent} \hspace{\algorithmicindent} $parent=n_l)$
                \State $n_l.child.append(\hat{n})$
            \EndFor
        \EndIf

        \For{each $k \in [1,2,\dots,B]$}
            \If{$n_l.child[k].V = 0$}
                \State $n_l \gets n_l.child[k]$ \textbf{break}
            \EndIf
        \EndFor
        
        \State \Return $n_l$
    \end{algorithmic}
\end{algorithm}

\begin{algorithm}[!t]
    \caption{Simulation}
    \label{algsim}
    \begin{algorithmic}[1]
        \Require Expanded node $n_l$, model for simulation $\phi_{simulation}$
        \Ensure Simulated trajectory $t_l$ and result $d_l$
        
        \State $t_l, d_l \sim \phi_{simulation}(n_l)$
        
        \State \Return $t_l, d_l$
    \end{algorithmic}
\end{algorithm}

\begin{algorithm}[!t]
    \caption{Backpropagation}
    \label{algback}
    \begin{algorithmic}[1]
        \Require Expanded node $n_l$, model for producing plausibility reward $\phi_{eval}$, simulated trajectory $t_l$, simulated result $d_l$, student chosen distractor $d$, plausibility reward weight $\alpha$
        
        \State match reward $r_1 \gets (d_l = d)$
        \State plausibility reward $r_2 \sim \phi_{eval}(t_l)$
        \State reward $r \gets r_1 + \alpha r_2$
        
        \While{$n_l$ has a parent node}
            \State $n_l.Q \gets n_l.Q + r$
            \State $n_l.V \gets n_l.V + 1$
            \State $n_l \gets n_l.parent$
        \EndWhile
    \end{algorithmic}
\end{algorithm}

\begin{algorithm}[!t]
    \caption{is\_fully\_expanded}
    \label{algfully}
    \begin{algorithmic}[1]
        \Require Node $n$
        \Ensure Whether this node is fully expanded
        
        \If{$n$ is a terminal node}
            \State \Return True
        \EndIf

        \For{$\hat{n} \in n.child$}
            \If{$\hat{n}.V = 0$}
                \State \Return False
            \EndIf
        \EndFor

        \State \Return True
    \end{algorithmic}
\end{algorithm}

\begin{algorithm}[!t]
    \caption{search\_match\_nodes}
    \label{algmatch}
    \begin{algorithmic}[1]
        \Require Current node $n$, student chosen distractor $d$, current list of matched terminal nodes $MNs$
        \Ensure Updated $MNs$
        
        \If{$n$ is a terminal node and $n.r = d$}
            \State $MNs.append(n)$
            \State \Return $MNs$
        \EndIf

        \For{$\hat{n} \in n.child$}
            \State $MNs \gets search\_match\_nodes(\hat{n},$
            \Statex \hspace{\algorithmicindent} $d, MNs)$
        \EndFor
        
        \State \Return $MNs$
    \end{algorithmic}
\end{algorithm}

\subsection{MCTS Process Details}
\begin{figure*}[htbp]
\centering
\includegraphics[width=0.95\linewidth]{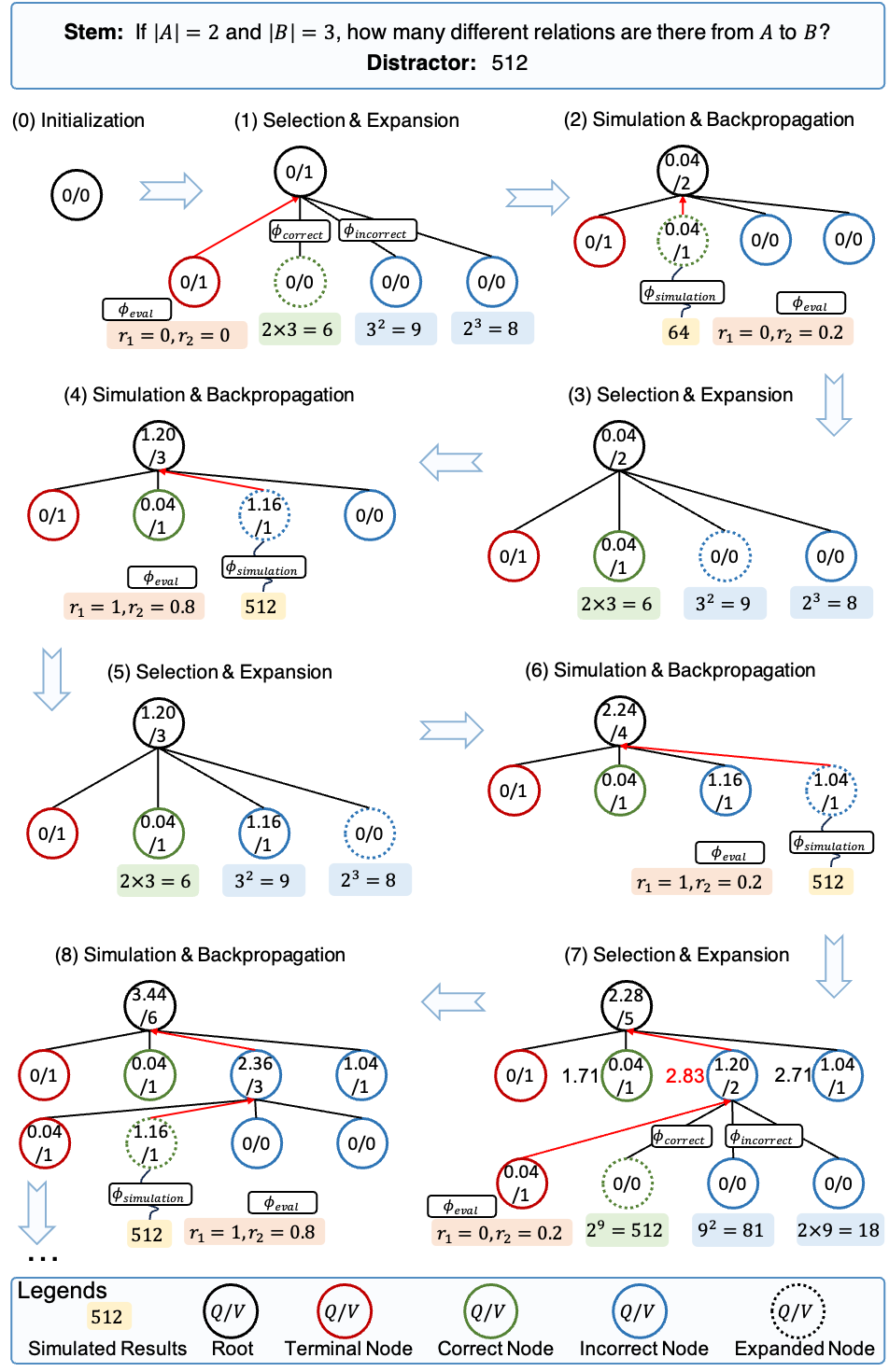} 
\caption{Illustration of the MCTS-based reasoning trajectory reconstruction process.}
\label{fpipeline}
\end{figure*}

We provide a detailed illustration of the MCTS-based reasoning trajectory reconstruction process in Figure \ref{fpipeline}. 
Starting from the root node, MCTS performs repeated cycles of selection, expansion, simulation, and backpropagation to explore plausible reasoning steps and recover the trajectory that best explains the student’s chosen distractor.
Below, we highlight three key design principles in this process.

First, \textbf{although a terminal node inherits the reasoning process and intermediate result of its parent node, it does not inherit the parent's reward and must be evaluated separately.}
A terminal node is evaluated at the moment when its parent node is expanded for the first time, \textit{i.e.}, when the terminal node is first generated, as illustrated in step (1) and (7) of Figure \ref{fpipeline}. 
At this point, the terminal node represents the decision to stop at the current step and treat the intermediate result as the final answer. 

Although it shares the same reasoning history as its parent, it cannot reuse the parent's reward, because the parent's reward is obtained through simulation that continues beyond the current step and reflects its \textit{future potential}. 
In contrast, the terminal node involves no further simulation, as it corresponds to a completed ``submit-answer'' state. 
Therefore, its reward must be computed directly from the terminal output (\textit{e.g.}, match and plausibility). 
After this one-time evaluation and backpropagation, the terminal node is never selected or expanded again, and its visit count remains fixed at $V=1$, functioning as an absorbing end state rather than a search frontier.

Second, \textbf{the model $\phi_{correct}$ continues reasoning from the student’s existing (possibly incorrect) trajectory, rather than regenerating a globally correct solution.}
Our goal is to reconstruct how the student might have reasoned, instead of overwriting earlier mistakes. 
Accordingly, when $\phi_{correct}$ is invoked, it does not restart the reasoning process.
Rather, it produces a logically coherent continuation that is conditioned on the existing partial reasoning---even if prior steps contain errors---thereby ``continuing correctly from a wrong premise'' while preserving the cognitive context of the misconception.

Third, \textbf{reaching the target distractor during simulation does not terminate the search; we continue exploring and later select the most plausible trajectory.}
During simulation, the search may reach the target distractor at an early stage and form a complete reasoning chain, as shown in step (4) and (6) of Figure \ref{fpipeline}. 
However, such trajectories may still be implausible or inconsistent with realistic student behavior. 
Therefore, we do not stop once a trajectory matches the distractor. 
Instead, MCTS continues exploring alternative paths; after the search completes, multiple trajectories may reach the same distractor, and we select the one with the highest plausibility score as the final recovered reasoning path.

\subsection{Scalability, Cost, and Computational Feasibility}
\label{appcost}
Although our framework introduces a structured multi-stage pipeline rather than a single-pass generator, its computational and deployment cost remains controllable in practice. 
First, the approach does not depend on commercial LLMs: as shown in Table \ref{tmain}, open-source backbones such as LLaMA-3-70B achieve over 70\% of the performance of the strongest commercial models (\textit{i.e.}, Claude-4-Sonnet). This enables cost-efficient deployment by substituting commercial APIs with locally hosted models when desired, without modifying the framework design.

Second, most computation occurs during offline prototype construction, which is inherently parallelizable across students, questions, and tasks. In addition to model-side parallelism, the framework also supports batched and asynchronous API execution, allowing multiple reasoning calls to be processed concurrently. At inference time, the online cost is lightweight: misconception prototypes are reused, and only a small number of search steps are executed per new question.

Third, the computational scale of each component is explicitly bounded. Each student contributes only a limited number of past QA records within a short temporal window, resulting in tens of records rather than hundreds in our datasets. Likewise, the number of extracted knowledge concepts per question is capped to 15 to avoid redundancy, which constrains both retrieval and reasoning complexity. These design choices ensure that the overall search space remains compact and computationally manageable.

Taken together, these properties demonstrate that the proposed framework is scalable, cost-aware, and feasible for practical deployment, while retaining the interpretability and diagnostic benefits enabled by explicit reasoning-based personalization.

\subsection{Error Attribution and Pipeline Traceability}
\label{apperror}
To ensure that errors in a multi-stage pipeline can be analyzed and localized rather than being treated as opaque failures, our system is designed with explicit stage-level traceability. For every run, we log each model invocation and store the outputs of all intermediate components, including concept extraction, trajectory reconstruction, misconception summarization, and distractor prediction. This allows the full reasoning chain to be replayed and inspected when an unexpected output occurs.

When an error is observed in the final distractor generation result, we can back-trace the pipeline to identify whether the failure originates from concept retrieval, reasoning reconstruction, prototype construction, or the final prediction stage. Each module can then be evaluated independently using its logged inputs and outputs, enabling targeted debugging rather than end-to-end trial-and-error adjustment. This design makes the system interpretable, diagnosable, and incrementally improvable, despite its multi-component structure.

\begin{figure}[t]
\centering
\includegraphics[width=\linewidth]{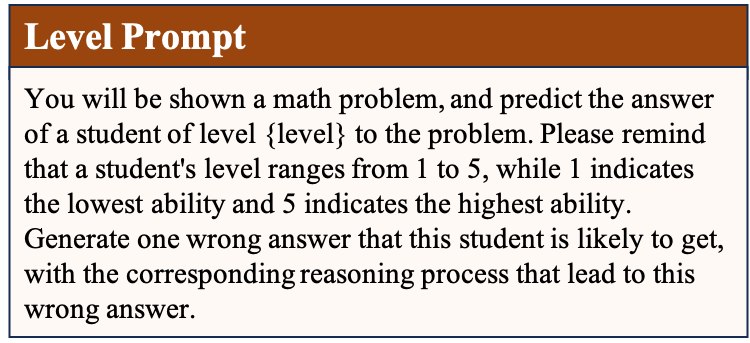} 
\caption{Details of the level prompt.}
\label{flevel}
\end{figure}

\section{More Experimental Settings and Results}
\label{aexp}
\subsection{Model Details}
\textbf{LLaMA-3-70B} \cite{llama} is a 70B multilingual language model by Meta AI, released in Dec 2024. The Instruct version is tuned for dialogue and achieves strong performance across open- and closed-source benchmarks. The tested API version is \textit{LLaMA-3-70B-Instruct}.

\noindent\textbf{DeepSeek-V3} \cite{liu2024deepseek} is an open-source language model developed by DeepSeek, supporting both English and Chinese. It is optimized for instruction-following tasks and achieves competitive performance on a range of benchmarks.

\noindent\textbf{Claude} \cite{claude} is Anthropic’s safety-focused language model designed for general-purpose tasks. The evaluated version is \textit{claude-sonnet-4-20250514}, known for strong reasoning and alignment.

\noindent\textbf{GPT-3.5} \cite{gpt35} is a widely used OpenAI model optimized for general natural language understanding and generation. The tested version is \textit{gpt-3.5-turbo-0125}.

\noindent\textbf{GPT-4o} \cite{achiam2023gpt} is OpenAI’s flagship model with advanced reasoning and multimodal support. The tested version is \textit{gpt-4o-2024-11-20}.

\subsection{Baseline Details}
We compare our method with six heuristic baselines that do not model student reasoning trajectories and two reasoning-aware baselines that attempt to reconstruct student thinking:
\begin{enumerate}
    \item \textit{Random}, which randomly selects a record from past QA records as a reference.
    
    \item \textit{Similarity}, which selects a record based on task statement similarities.

    \item \textit{Level}, which estimates a student's ability level based on the accuracy of their past QA records. Specifically, we first compute the accuracy of a student's past QA records and then normalize it to a range of 1 to 5, where 5 represents the highest cognitive level. As mentioned in~\citet{level}, this relative cognitive level simulation approach yields better results. We incorporate this level into the prompt to indicate the student's proficiency, enabling the model to infer student misconception accordingly. The prompt is shown in Figure \ref{flevel}.

\begin{figure}[t]
\centering
\includegraphics[width=\linewidth]{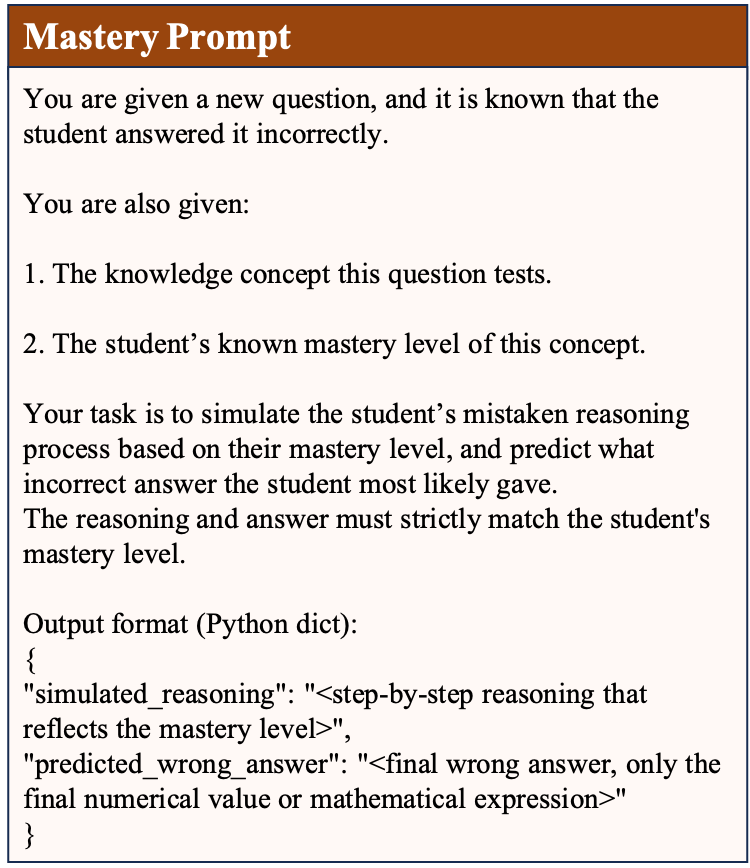} 
\caption{Details of the mastery prompt.}
\label{fmastery}
\end{figure}

    \item \textit{Level+Random}, which incorporates \textit{Random} and \textit{Level}. We add the randomly selected past QA record into the \textit{Level} prompt.
    
    \item \textit{Level+Similarity}, which incorporates \textit{Random} and \textit{Similarity}. We add the similarity-based retrieved past QA record into the \textit{Level} prompt.

\begin{figure}[t]
\centering
\includegraphics[width=\linewidth]{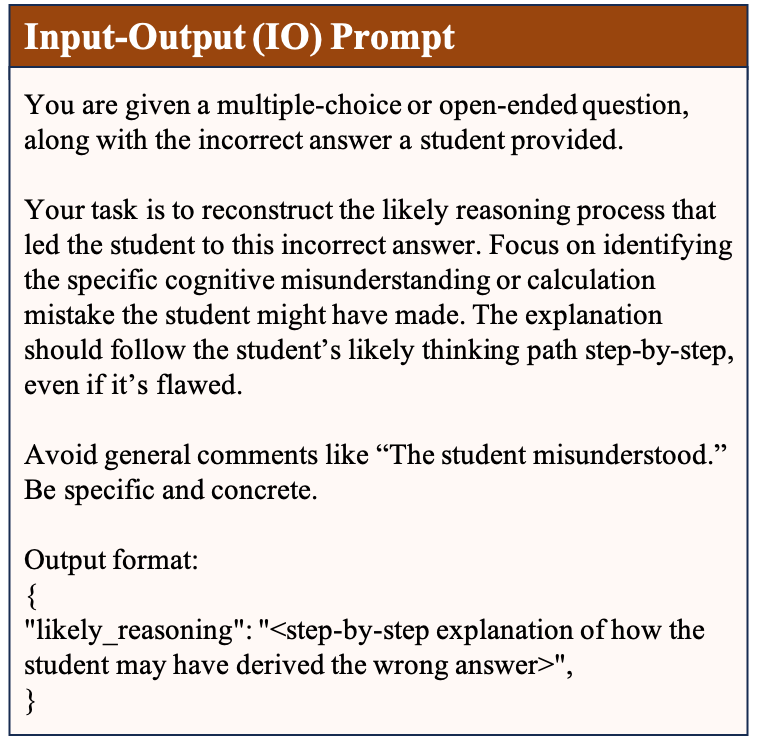} 
\caption{Details of the IO prompt.}
\label{fio}
\end{figure}

\begin{figure}[t]
\centering
\includegraphics[width=\linewidth]{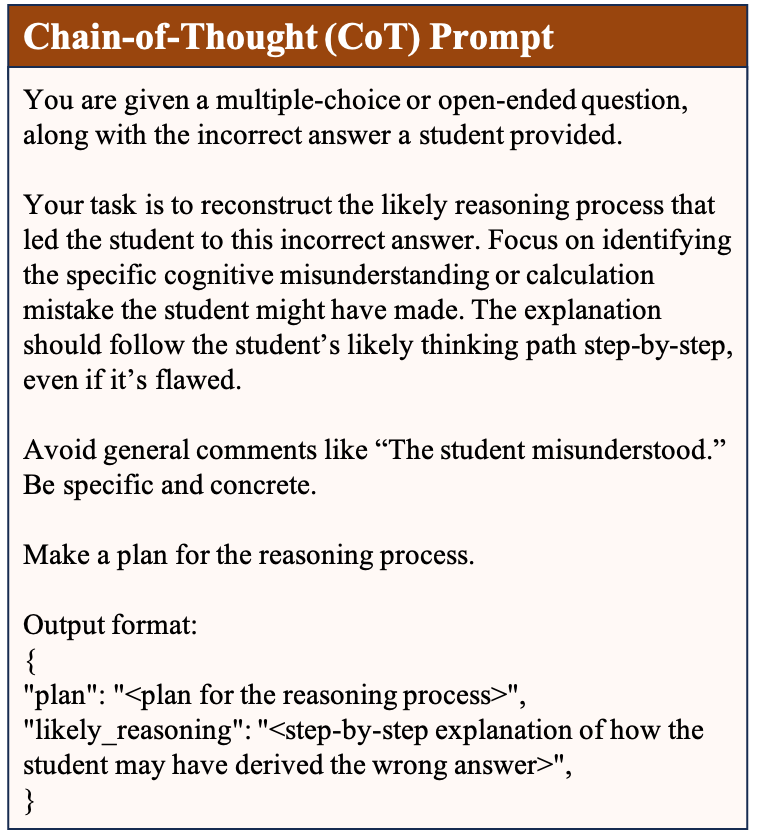} 
\vspace{-1em}
\caption{Details of the CoT prompt.}
\label{fcot}
\vspace{-1em}
\end{figure}
    
    \item \textit{Mastery} \cite{studentsimulation}, which models the student’s mastery over each extracted knowledge concept without inferring the underlying reasoning trajectory. As a result, it can only capture shallow information—such as whether the student understands a concept well or poorly—while failing to identify how the student is likely to make mistakes. The prompt is shown in Figure \ref{fmastery}.
    
    \item \textit{IO} \cite{tot}, which requires models to directly output the reasoning trajectories. This prompt is shown in Figure \ref{fio}.

\begin{figure*}[t]
\centering
\includegraphics[width=\linewidth]{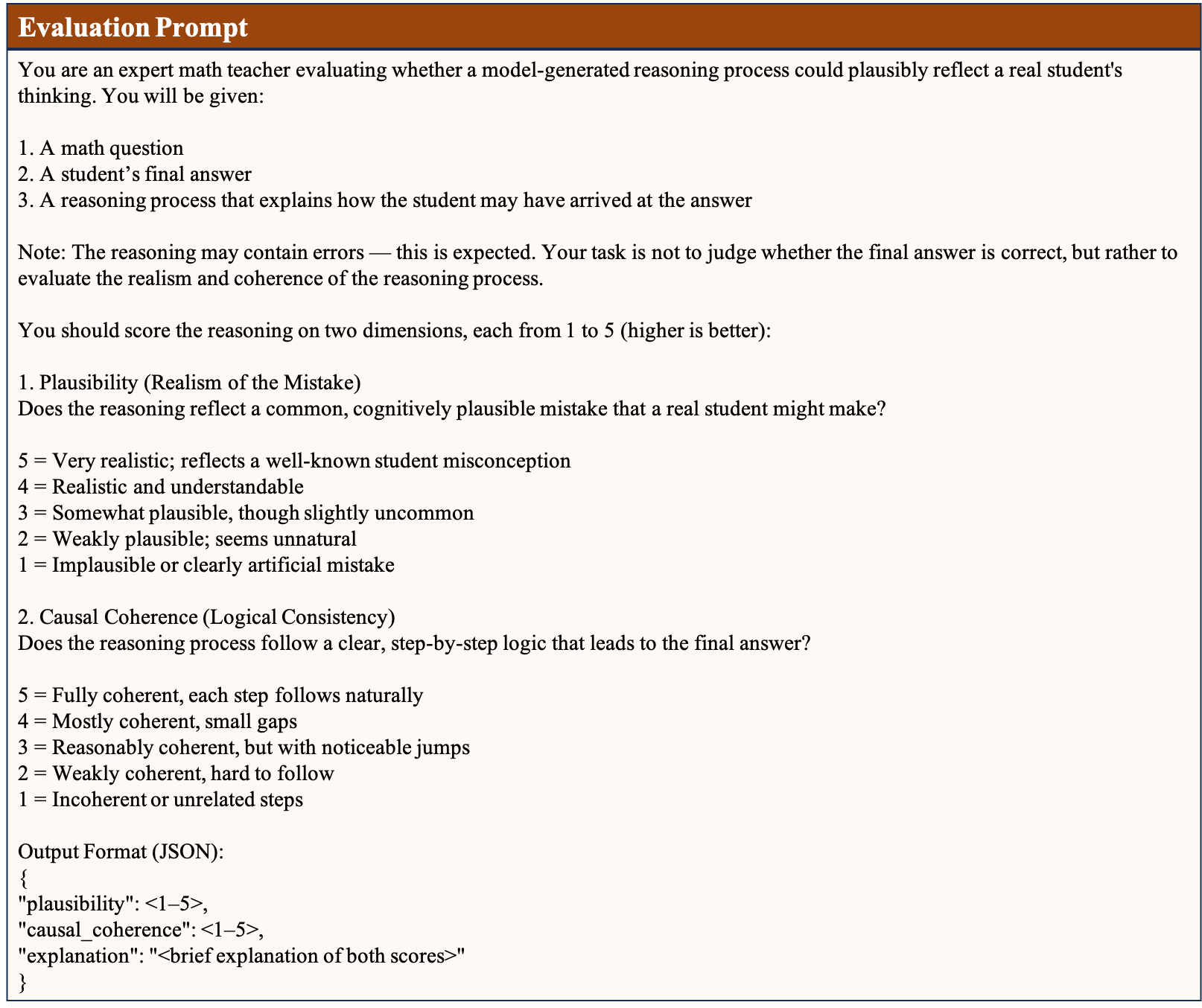} 
\caption{Details of the prompt for producing Plaus and Coh scores.}
\label{feval}
\end{figure*}

    \item \textit{CoT}~\cite{cot}, which introduces a chain of thoughts that connects the input to the output, with each thought forming a coherent language sequence that acts as a meaningful intermediate step toward solving the problem. This prompt is shown in Figure \ref{fcot}.
\end{enumerate}

\subsection{Metric Details}
We propose two LLM-based metrics to evaluate the quality of generated reasoning trajectories. Specifically, we use the o3-mini model (API version: \textit{o3-mini-2025-01-31}), which is an advanced open-source model, to ensure fair and consistent evaluation. Detailed prompt designs for these metrics are illustrated in Figure~\ref{feval}.

\subsection{Ablation Study on Hyperparameter Choices}
We conduct a sensitivity analysis to investigate how key parameters affect model performance. Specifically, we examine the impact of the MCTS tree depth $D$, the maximum number of child nodes per reasoning step $B$, and the weight $\alpha$ used to balance plausibility and match scores. The results are reported in Table~\ref{td}, \ref{tb} and \ref{talpha}.

We find that varying the tree depth $D$ has limited impact on performance. In practice, due to the fixed number of search iterations, the tree seldom grows to its maximum depth. However, setting $D$ too small may harm performance in complex domains like Discrete Math, where reasoning often involves multiple intermediate steps.

For the child budget $B$, we observe that $B=3$ yields the best results. A smaller value restricts the model’s ability to explore diverse plausible mistakes, while a larger value spreads the search budget too thin across many shallow branches, limiting deeper exploration of each reasoning trajectory.

The coefficient $\alpha$ controls the weight of plausibility scores in the search. We find that $\alpha=0.2$ achieves the best trade-off. When $\alpha$ is too small, plausibility receives little weight during node evaluation. As a result, in the early stages of search—when matched distractors are rarely encountered—the overall reward remains low across nodes, making it difficult to differentiate promising search paths. In this case, the search lacks meaningful guidance and becomes ineffective. Conversely, when $\alpha$ is too large, plausibility dominates over match scores, leading the model to prioritize fluent or logically sound paths that may not align with the student’s actual misconceptions.

\subsection{Generalization to Group-level Distractor Generation}
We further evaluate the generalizability of our framework in the group-level distractor generation setting. We compare it with three established baselines—D-GEN \cite{dgen}, O\&A \cite{oanda}, and DiVERT \cite{divert}—where O\&A and DiVERT are trained on the Eedi dataset. All baseline methods generate a fixed set of three distractors per question.

\begin{table}[t]
\centering
\resizebox{0.5\linewidth}{!}{
\begin{tabular}{c|cc}
\toprule
$D$ & Eedi\_100 & Discrete\_40 \\
\midrule\midrule
3 & 22.1 & 27.1\\
4 & 22.2 & 28.2\\
5 & 22.1 & 28.6 \\
6 & 22.0 & 28.6\\
7 & 21.8 & 28.8\\
 \bottomrule
\end{tabular}
}
\caption{Ablation results (Acc) on MCTS tree depth $D$.}
\label{td}
\end{table}

\begin{table}[t]
\centering
\resizebox{0.5\linewidth}{!}{
\begin{tabular}{c|cc}
\toprule
$B$ & Eedi\_100 & Discrete\_40 \\
\midrule\midrule
2 & 20.7 & 26.7\\
3 & 22.1 & 28.6 \\
4 & 21.0 & 29.0 \\
5 & 20.5 & 26.1 \\
 \bottomrule
\end{tabular}
}
\caption{Ablation results (Acc) on maximum number of child nodes $B$.}
\label{tb}
\end{table}

\begin{table}[t]
\centering
\resizebox{0.5\linewidth}{!}{
\begin{tabular}{c|cc}
\toprule
$\alpha$ & Eedi\_100 & Discrete\_40 \\
\midrule\midrule
0 & 16.8 & 24.5\\
0.1 & 21.1 & 28.2 \\
0.2 & 22.1 & 28.6 \\
0.3 & 22.1&  27.8\\
0.5 & 21.7 & 25.9\\
1.0 & 20.3 & 23.7 \\
 \bottomrule
\end{tabular}
}
\caption{Ablation results (Acc) on plausibility score weight $\alpha$.}
\label{talpha}
\end{table}

In contrast, our method first operates at the personalized level. For each student, we use our framework to generate a personalized distractor for every question that matches the student’s ability level (for example, students in \texttt{Eedi\_100} answer only elementary-level questions, while students in \texttt{Discrete\_40} are evaluated on discrete-math questions). This produces a pool of individualized distractors reflecting diverse student-specific misconceptions. We then aggregate all generated distractors across students on a per-question basis, compute their empirical frequency, and select the top three most frequent distractors as the final group-level set. This aggregation reflects the intuition that misconceptions frequently observed across personalized cases correspond to shared misconception patterns at the group level.

We evaluate this process on questions from \texttt{Eedi\_100} and \texttt{Discrete\_40}, and additionally include two held-out datasets—Elementary Math from MMLU \cite{mmlu} and Discrete Math from CEval \cite{ceval}—to assess out-of-domain generalization. Performance is measured using Recall, which quantifies how well the generated distractors cover the actual distractors selected by students in real QA records (higher Recall indicates stronger coverage of real misconception behaviors).

Results in Table \ref{tgeneral_dg} show that baseline models perform well on in-domain questions but degrade substantially when evaluated on the held-out datasets. By contrast, our method maintains consistently strong performance across both in-domain and out-of-domain settings, demonstrating robustness and superior generalization. These findings suggest that personalization not only benefits individual diagnosis, but also provides an effective pathway for deriving high-quality group-level distractors from aggregated personalized misconceptions.

\begin{table}[t]
\centering
\resizebox{\linewidth}{!}{
\begin{tabular}{c|cc|cc}
\toprule
\multirow{2}[2]{*}{\textbf{Methods}} & \multicolumn{2}{c|}{Elementary Math} & \multicolumn{2}{c}{Discrete Math} \\
\cmidrule{2-5}
 & Eedi\_100 & MMLU & Discrete\_40 & CEval \\
 \midrule\midrule
 \rowcolor{gray!20} \multicolumn{5}{c}{\textit{Group-level Distractor Generation Baselines}} \\
 D-GEN & 20.48 & 18.28 & 28.57 & 28.89 \\
 O\&A & 26.63 & 22.22 & 20 & 18.89 \\
 DiVERT & 31.66 & 25.09 & 20.71 & 20 \\
 \midrule
\rowcolor{gray!20} \multicolumn{5}{c}{\textit{Our Method}} \\
LLaMA-3-70B & 22.16 & 30.29 & 28.57 & 36.67 \\
Deepseek-V3 & 31.1 & 34.95 & 36.43 & 40 \\
Claude-4-Sonnet & 28.31 & 32.44 & 34.29 & 45.56 \\
GPT-3.5-turbo & 24.39 & 27.96 & 30.71 & 44.44 \\
GPT-4o & 27.37 & 30.29 & 29.29 & 45.56 \\

 \bottomrule
\end{tabular}
}
\caption{Performance comparison on group-level distractor generation across four datasets.}
\label{tgeneral_dg}
\end{table}

\begin{table*}[t]
\centering
\resizebox{0.75\linewidth}{!}{
\begin{tabular}{cc|cc|cc|cc|cc}
\toprule
\multicolumn{2}{c|}{LLaMA-3-70B} & \multicolumn{2}{c|}{Deepseek-V3} & \multicolumn{2}{c|}{Claude-4-Sonnet} & \multicolumn{2}{c|}{GPT-3.5-turbo} & \multicolumn{2}{c}{GPT-4o} \\
 Precision & Recall & Precision & Recall & Precision & Recall & Precision & Recall & Precision & Recall\\
 \midrule\midrule
 0.7 & 0.76 & 0.74 & 0.75 & 0.65 & 0.83 & 0.71 & 0.77 & 0.72 & 0.75 \\
 
 \bottomrule
\end{tabular}
}
\caption{Evaluation of concept extraction accuracy of our method on the \texttt{Discrete\_40} dataset.}
\label{tkc}
\end{table*}

\begin{table*}[t]
\centering
\resizebox{0.7\linewidth}{!}{
\begin{tabular}{c|ccccc}
\toprule
Methods & LLaMA-3-70B & Deepseek-V3 & Claude-4-Sonnet & GPT-3.5-turbo & GPT-4o \\
 \midrule\midrule
 CoT & 0.33 & 0.51 & 0.62 & 0.39 & 0.48 \\
 MCTS & 0.46 & 0.6 & 0.74 & 0.5 & 0.63 \\
 
 \bottomrule
\end{tabular}
}
\caption{Mechanism match accuracy of MCTS vs. CoT across LLM backbones.}
\label{tmma}
\end{table*}

\subsection{Analysis of Lower Acc Scores in \texttt{CP\_39} and \texttt{OS\_124}}
We observe that the Acc values for Compiler Principles (\texttt{CP\_39}) and Operating Systems (\texttt{OS\_124}) are noticeably lower than those in the other subjects in Table \ref{twhole}.
This does not primarily result from poor generation quality, but rather from characteristics of these domains. 
First, although we have removed text-heavy options and retained only distractors consisting of numbers, expressions, or code, the remaining distractors in \texttt{CP\_39} and \texttt{OS\_124} are still substantially more complex than those in other subjects, often involving long or highly structured expressions. 
As a result, different students may express similar misconceptions through different but semantically related forms, making matching to the specific distractor previously selected by a student much more difficult. 
Second, misconceptions in these theoretical, concept-intensive subjects tend to be more diverse and less concentrated than in mathematics or programming tasks, where students’ errors are often clustered around a few recurring patterns. 
Consequently, our model may generate a plausible distractor that accurately reflects a student’s reasoning error, but it does not exactly coincide with the particular option chosen in the dataset, leading to lower Acc despite reasonable Plaus and Coh scores. 
These observations suggest that the drop in Acc mainly reflects the strictness of the exact-match metric under domains with heterogeneous misconception expressions, rather than a degradation in the model’s ability to model student reasoning.

\subsection{Evaluation of Intermediate Pipeline Outputs}
Since our pipeline is relatively large and involves multiple interdependent components, we further evaluate the reliability of its intermediate outputs. 
\subsubsection{Evaluation of Knowledge Concept Extraction}
In particular, we assess the accuracy of the knowledge concept extraction process on the \texttt{Discrete\_40} dataset, where each question is annotated with its corresponding concepts by the online learning platform. 
We treat these platform-provided annotations as ground truth and measure the precision and recall of the concepts extracted by our method. 
As shown in Table \ref{tkc}, our approach achieves consistently high precision and recall, indicating that the extracted concepts closely match the true conceptual structure of the questions and provide a reliable basis for subsequent reasoning reconstruction and distractor generation.

\subsubsection{Evaluation of MCTS-reconstructed Reasoning}
Empirically, we find that the MCTS-based reconstruction captures student error patterns to a meaningful extent. Specifically, we conduct additional validation for the MCTS reconstruction from both mechanistic alignment and counterfactual behavioral consistency perspectives.

\textbf{Mechanistic alignment consistency.} Importantly, the platform-sourced portion of the \texttt{Student\_1361} dataset consists of high-quality MCQs authored by domain experts, and a subset of these questions includes expert-written error rationales for each distractor (\textit{i.e.}, explicit descriptions of the underlying misconception or erroneous rule). Leveraging this resource, we randomly identify 30 questions in the \texttt{Discrete\_40} subset that contain usable expert rationales, covering 90 distractors in total.

For each distractor, we run MCTS and CoT (as baseline) to obtain the highest-reward erroneous reasoning trajectory. Independent domain experts, who are blind to the source of the trajectories, then judge whether the key error mechanism implied by the reconstructed trajectory matches the expert-provided rationale (allowing differences in wording or intermediate steps as long as the core misconception is consistent). We report Mechanism Match Accuracy as the evaluation metric in Table \ref{tmma}. Across different LLM backbones, MCTS-based reconstruction achieves better agreement with expert rationales than CoT-based baselines, indicating that our method reliably recovers the typical error mechanisms underlying distractors.

\textbf{Counterfactual behavioral consistency.}
To further provide behavioral evidence at the individual level, we design a counterfactual variant experiment. The core hypothesis is: if MCTS correctly reconstructs the erroneous reasoning process that led a student to select a particular distractor, then under an isomorphic problem variant (same reasoning structure with perturbed numerical values), the distractor recomputed from the same erroneous reasoning process should remain attractive to that student and thus be selected again with high probability.

Concretely, we sample 5 students from \texttt{Discrete\_40} and select 5 historical incorrect questions per student (25 total), focusing on formula-driven problems with stable reasoning structures (\textit{e.g.}, Euler’s formula, counting rules). For each original question, we first use MCTS with the best underlying model, Claude-4-Sonnet, to reconstruct the erroneous reasoning trajectory corresponding to each distractor. We then create isomorphic variants by perturbing only the numerical values while preserving the underlying reasoning structure. For each variant, we recompute distractors by executing the same reconstructed erroneous reasoning paths, and combine them with the correct answer to form the new option set. Students then answer these variant questions in randomized order to minimize memory effects.

We use Behavioral Match@1 as the primary metric: a match is counted if the student again selects the distractor derived from their past QA records. The results show strong behavioral consistency, with Behavioral Match@1 reaching 0.68. This provides additional evidence that the MCTS-reconstructed reasoning captures student-specific error tendencies in a behaviorally meaningful way.

\subsection{Bad Case Analysis}
Overall, our framework exhibits stable performance across most students and question types, and is generally able to recover meaningful misconception patterns from historical QA records. Nevertheless, we also observe occasional failure cases that reflect the inherent difficulty of distinguishing stable reasoning tendencies from incidental errors in student behavior.

In particular, some errors arise when a student’s past records contain isolated, non-systematic mistakes—such as accidental mis-clicks, arithmetic slips, or atypical one-off reasoning attempts—that do not reflect the student’s true misconception but are still reconstructed as plausible trajectories during MCTS. We note that this phenomenon is not specific to our approach: because such incidental mistakes are embedded in the raw QA data, they can influence any method that learns or infers patterns from historical records, including the baseline systems. In all such cases, the risk stems from the data itself rather than from the modeling mechanism \cite{ood1, ood2}.

These cases tend to occur when the available history is sparse and heterogeneous, so that a small number of atypical mistakes exert disproportionate influence on the recovered prototype. We explored whether expanding the retrieval scope to include more past records could mitigate this issue, but found that retrieving additional tasks may also introduce irrelevant or weakly related reasoning traces, enlarging the prompt and increasing semantic noise, which can in turn degrade prediction quality. This suggests that the challenge lies not in data quantity alone, but in the fundamental difficulty of separating incidental noise from cognitively meaningful evidence—a challenge faced broadly across personalization-based modeling approaches.

It is worth noting that such edge cases are relatively rare and do not affect the overall stability of our framework, whose primary focus is to capture consistent, recurring misconception tendencies rather than isolated accidental behaviors. Fully resolving this phenomenon would require more advanced mechanisms for noise-aware misconception modeling, such as temporal weighting, confidence filtering, or selective prototype updating.

Finally, we note that similar challenges have also been reported in other domains—such as recommender systems—where spurious user behaviors or accidental interactions can bias profile construction \cite{causal1, sigma, causal2}. In those settings, handling such noise often motivates separate lines of research (\textit{e.g.}, debiasing, counterfactual modeling, or causal filtering \cite{modelgpt, mnemis, causal3}) rather than being solved within a single pipeline. We view our observation in educational data as part of this broader challenge, and consider it a promising direction for future independent investigation.

\stopcontents
\end{document}